\pdfoutput=1
\documentclass[11pt]{article}

\usepackage{acl}

\usepackage{times}
\usepackage{latexsym}

\usepackage[T1]{fontenc}
\usepackage[utf8]{inputenc}
\usepackage{amssymb}
\usepackage{array}
\usepackage{lipsum}  
\usepackage{multirow}

\usepackage{microtype}
\usepackage{graphicx}
\usepackage{booktabs}
\usepackage{subcaption}
\usepackage{amsmath}
\usepackage[export]{adjustbox}

\DeclareMathOperator*{\argmax}{arg\,max}

\newcolumntype{L}[1]{>{\raggedright\let\newline\\\arraybackslash\hspace{0pt}}m{#1}}
\newcolumntype{C}[1]{>{\centering\let\newline\\\arraybackslash\hspace{0pt}}m{#1}}
\newcolumntype{R}[1]{>{\raggedleft\let\newline\\\arraybackslash\hspace{0pt}}m{#1}}

\usepackage{listings}

\newcommand{\ex}[1]{``\emph{#1}''}
\newcommand*{\affaddr}[1]{#1} % No op here. Customize it for different styles.
\newcommand*{\affmark}[1][*]{\textsuperscript{#1}}
\newcommand*{\email}[1]{\texttt{#1}}

\title{Generating Coherent Sequences of Visual Illustrations for \\Real-World Manual Tasks}
\author{João Bordalo\affmark[1]$\:\:\: \textbf{Vasco Ramos}\affmark[1]$ \:\:\: \textbf{Rodrigo Valério}\affmark[1]\:\:\: \textbf{Diogo Glória-Silva}\affmark[1]\\
\:\:\: \textbf{Yonatan Bitton}\affmark[2] \:\:\:\textbf{Michal Yarom}\affmark[2] \:\:\:\textbf{Idan Szpektor}\affmark[2] \:\:\:\textbf{Joao Magalhaes}\affmark[1] \:\:\:\\
\affaddr{\affmark[1]Universidade NOVA de Lisboa}\\
\affaddr{\affmark[2]Google Research}\\
\email{jmag@fct.unl.pt}, 
\email{szpektor@google.com}\\
}

\begin{document}
\maketitle
\begin{abstract}
Multistep instructions, such as recipes and how-to guides, greatly benefit from visual aids, such as a series of images that accompany the instruction steps. While Large Language Models (LLMs) have become adept at generating coherent textual steps, Large Vision/Language Models (LVLMs) are less capable of generating accompanying image sequences. The most challenging aspect is that each generated image needs to adhere to the relevant textual step instruction, as well as be visually consistent with earlier images in the sequence.
To address this problem, we propose an approach for generating consistent image sequences, which integrates a Latent Diffusion Model (LDM) with an LLM to transform the sequence into a caption to maintain the semantic coherence of the sequence. In addition, to maintain the visual coherence of the image sequence, we introduce a copy mechanism to initialise reverse diffusion processes with a latent vector iteration from a previously generated image from a relevant step.
Both strategies will condition the reverse diffusion process on the sequence of instruction steps and tie the contents of the current image to previous instruction steps and corresponding images.
Experiments show that the proposed approach is preferred by humans in 46.6\% of the cases against 26.6\% for the second best method. In addition, automatic metrics showed that the proposed method maintains semantic coherence and visual consistency across the sequence of visual illustrations.
\end{abstract}

\section{Introduction}
When humans undertake a task with numerous intricate steps, merely reading a step description is limiting, leaving the user to imagine and infer some of the more nuanced details~\cite{wizard_of_tasks}. Complementing the textual step instructions with images enhances the user experience by better communicating and representing the text semantics and ideas~\cite{serafini2014reading}.

\begin{figure}[t]
    \centering
    \includegraphics[width=\linewidth]{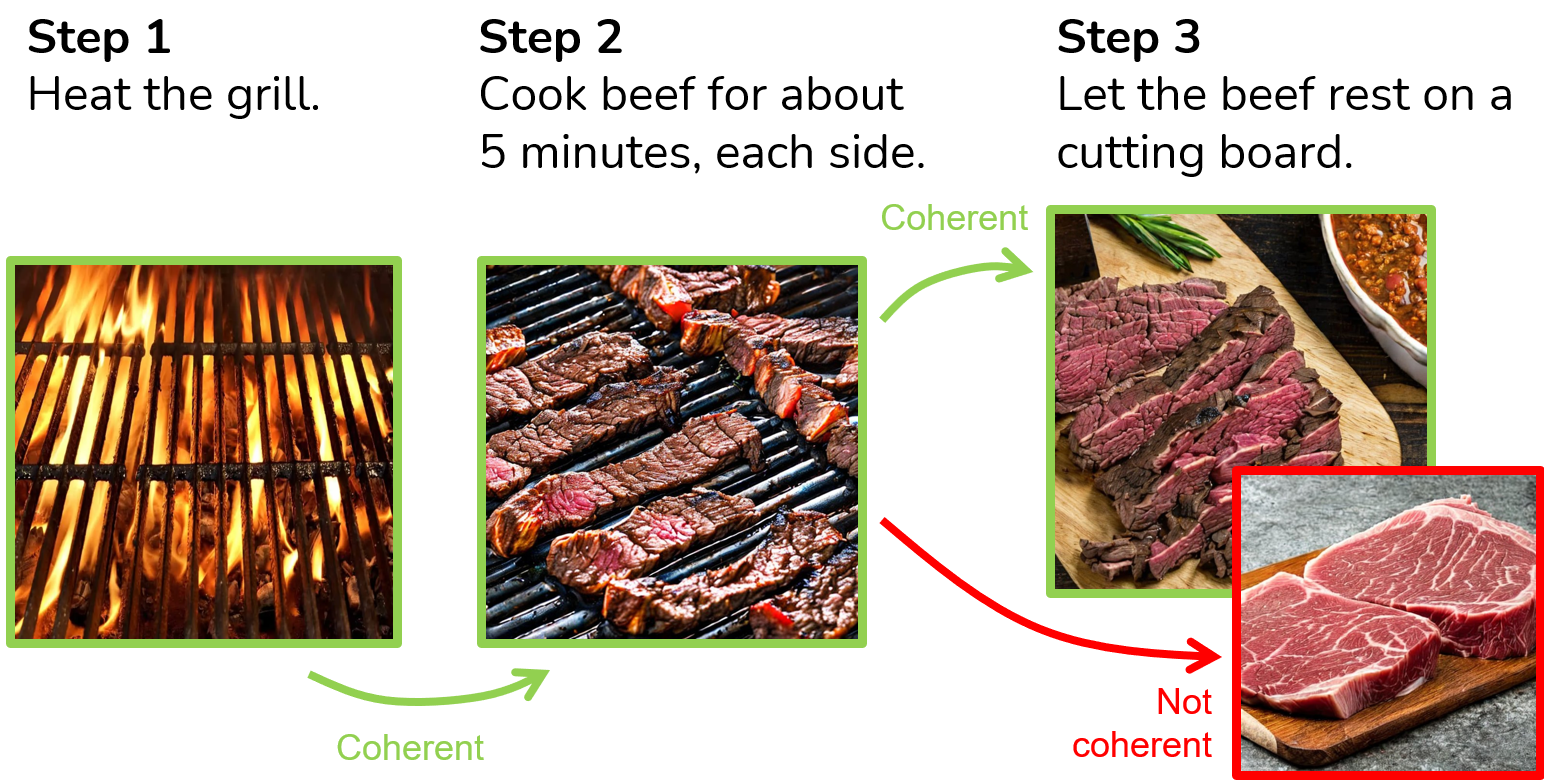}
    \caption{The properties of the elements in illustrations should remain coherent throughout the whole sequence.}
    \label{fig:beef-example}
\end{figure}

Although prompt-based image generation has advanced significantly \cite{betker2023improving,stable-diffusion,imagen}, state-of-the-art (SOTA) models such as Latent Diffusion Models (LDMs)~\cite{stable-diffusion} still struggle when generating image sequences to accompany textual instruction steps~\cite{lu2023multimodal}. The challenge lies in effectively combining two key aspects: (a) accurately portraying the actions outlined in the step instructions, and (b) ensuring coherence between successive images to avoid confusing the user. 
Existing storytelling approaches~\cite{ACM-VSG, AR-LDM, MakeStory} operate mostly on linear storytelling and use synthetic cartoon datasets with explicit sequence information, i.e., the textual prompts describe the images appropriately and have no implicit co-references. 
These aspects limit the applicability of existing methods to real-world scenarios (Figure~\ref{fig:beef-example}), where there is a lack of informative prompts accompanying images, and dependencies between prompts are not necessarily linear.

In this paper, we explore the generation of image sequences within two domains: recipe instructions, and Do It Yourself (DIY) guides, both showing increasing online consumption~\cite{bausch2021covid,social-media-for-recipes,sarpong2020yourself,quader2022central}. In these domains, accuracy and coherence are of utmost importance to ensure that the result of all manual actions is correct, and that the user is correctly guided to the target output, Figure~\ref{fig:beef-example}.
These domains contain (i) complex sequential manual tasks of detailed actions, (ii) coherence requirements for the images accompanying the sequence step descriptions, and (iii) a non-linear sequential structure, where steps may be related to earlier steps--not necessarily the previous step.

To tackle these challenges, we propose to extend Latent Diffusion Models~\cite{stable-diffusion}, with an LLM decoder to semantically condition the reverse diffusion process in the sequence of steps and a copy mechanism to select the best LDM initialisation.
The image generation process is conditioned on the current step and the previous steps, to increase semantic coherence. In addition, our method initializes the reverse diffusion process with a latent vector iteration copied from a previous generation process to ensure the visual coherence of the generated image.
Through this dual attendance to past textual and visual items in the sequence, we aim to achieve \emph{semantic coherence}, which pertains to the presence and persistence of objects in consecutive images, and \emph{visual coherence}, which aims to ensure the consistency of backgrounds and visual object properties across successive images.

Extensive automatic and manual evaluations confirmed that our model outperforms strong baselines in terms of the overall quality of the generated sequence of illustrations in the cooking and DIY domains.

\section{Related Work}
Methods to generate sequences of images, conditioned on textual input, have been explored in the story visualization and story continuation tasks.
Story Visualization aims at generating a coherent sequence of images, based on a multi-sentence paragraph or a series of captions forming a narrative~\cite{storygan, AR-LDM, StoryDALL-E}. 
Story Continuation is a variant of Story Visualization, in which the generated sequence is initiated by a source image.
In both tasks, generating every image independently of other images in the sequence leads to low visual coherence.

Several works addressed these tasks. \citet{AR-LDM} proposed AR-LDM, a method to tackle Story Visualization and Continuation using a history-aware autoregressive latent diffusion model~\cite{stable-diffusion}, which encodes the history of caption-image pairs into a multimodal representation that guides the LDM denoising process.
Despite the intuitive idea, the computational complexity of the conditioning network makes this approach too costly. Additionally, AR-LDM still shows room for improvement in terms of coherence.
\citet{ACM-VSG} noted that AR-LDM conditions the current generation on all historic frames and captions equally, despite not all frames being similarly related. To tackle this limitation, they proposed ACM-VSG, a method that selectively adopts historical text-image data for the generation of the new image.
The adaptive encoder automatically finds the relevant historical text-image pairs via CLIP similarity. A key difference between AR-LDM and ACM-VSG is the computational cost: while AR-LDM fine-tunes CLIP, BLIP, and the LDM, ACM-VSG trains only the cross-attention module, at a much lower cost.

LDMs use a cross-attention layer to condition the image denoising U-Net on an input text. \citet{rahman2022makeastory} used the full history of U-net latent vectors from all segments of the sequence, averaging these historic latent vectors in a cross-attention layer that is merged with the existing one in the LDM pipeline. In this method, image generation is conditioned on the segment text and on the entire set of latent vectors, with limited awareness of visual coherence of segments.

The above approaches focus on two synthetic cartoon datasets~\cite{Pororo,FlintstonesSV}, with limited characters and scenes. Among these,~\cite{AR-LDM} shows limited performance when applied to real-world sequences, and~\cite{ACM-VSG,MakeStory} do not evaluate their solutions in a real-world scenario. Additionally, these datasets have textual descriptions of the associated images, which are rarely available for real-world complex tasks -- the focus of this paper.

Finally, it is interesting to note the conceptual relation to the visual storytelling problem introduced by~\cite{huang-etal-2016-visual}, which aims to create a text story from a sequence of images~\cite{, Wang_Wei_Li_Zhang_Huang_2020,hsu-etal-2021-plot}. While, conceptually similar, this task can be seen as the inverse of the problem addressed in the current paper.

\section{Illustrating Real-World Manual Tasks}
\label{sec:multimodal-complex-manual-tasks}
We consider a set of manual tasks $\mathcal{D}$, where each task $TS \in \mathcal{D}$ is composed of a sequence of $n$ step-by-step instructions, $TS = \{ (s_1, v_1), ..., (s_n, v_n) \}$. A task step $(s_i, v_i)$, consists of a natural language instruction $s_i$, and its corresponding visual instruction $v_i$. 

Given the sequence of steps $\{s_1, \ldots, s_{n} \}$, our goal is to generate a sequence of images $\{v_1, \ldots, v_{n} \}$, in which $v_i$ visually represents step $s_i$. 
A step $s_i$ may be dependent on any number of previous steps, in a non-linear sequential structure~\cite{donatelli-etal-2021-aligning}. 
To generate each image accurately, the model needs to condition its output not only on $s_i$ but also on previous steps $\{s_1, \ldots, s_{i-1}\}$; this way, context is preserved even when steps are ambiguous or lack information, e.g. \ex{Add two eggs and mix}.
In addition to previous step instructions, we also need to condition on previously generated images, $\{v_1, ..., v_{i-1}\}$, to maintain the visual aspects that are only introduced in the images, such as object properties and background artefacts not mentioned in the text.

\section{Sequential Latent Diffusion Model}
The latent diffusion model proposed by \citet{stable-diffusion} transforms the diffusion process into a low-dimensional latent space through an encoder $z=E(v)$ and recovers the real image with a decoder $v=D(z)$. 
The complete model is also conditioned on an input $y$ by augmenting the U-Net backbone with a cross-attention layer to support the encoded input $\tau_\theta(y)$. The conditional LDM is learned with the loss,
\begin{equation}
    \mathcal{L}_{LDM} = \mathbb{E}_{E(v), y,\epsilon, t} \Bigl[ \| \epsilon - \epsilon_\theta(z_t,t, \tau_\theta(y)) \| ^2_2 \Bigr],
    \label{eq:loss_ldm}
\end{equation}
where $\epsilon \sim \mathcal{N}(0,1)$, $t$ is the denoising iteration, and the conditioning encoder $\tau_{\theta}(y)$ uses the entire set of tokens in $y$ to condition the U-Net denoising process in the LDM backbone.
Eq.~\ref{eq:loss_ldm} evidences how the conditional LDM is designed to generate one image $v$ at a time. More relevant to our problem is the fact that the latent vectors $z_t$ are independent across different image generations, since the reverse diffusion process iterates from $T$ to 1 starting with a new random seed $z_T$ for every new image generation.

\subsection{Sequence Context Decoder}
\label{subsub:task-context}
Generally, textual step descriptions describe what the user should do at a specific step of their manual task. These descriptions do not make accurate captions of the accompanying step images as they often contain information that is not visually representable, such as temporal information, \textit{"Cook for 10 minutes"}, or multiple actions, \textit{"Chop the rosemary, dice the carrots, and peel the cucumber."}
Additionally, it is also common for steps to not be self-contained, as they depend on the previous step descriptions for context.

To overcome this, for each step $s_i$, we use a decoder-only model, $\varphi$, which we call Sequence Context Decoder, to transform the step and its context into a visual caption $c_i$ which describes the contents of the image $v_i$. To ensure the generated captions are contextually relevant, we adopt a middle-ground approach and consider the target step $s_i$ and a context window of $w$ steps. Formally, we define the decoder
\begin{equation}
    c_i = \varphi(s_i, \{ s_{i-1}, \ldots s_{i-w}\})
\end{equation}
to generate a contextual caption $c_i$ from its step description $s_i$ and context.

The decoder $\varphi(\cdot)$ is trained similarly to an image caption generator, but instead of receiving images as input, it receives the step and its context. 
By training the model to output image captions for the original images that we are trying to replicate, the model learns to generate texts that are more appropriate as image generation prompts.
The objective now is learning how to generate better image generation prompts, instead of training the image generation module.

\begin{figure}[t]
    \centering
    \includegraphics[width=\linewidth]{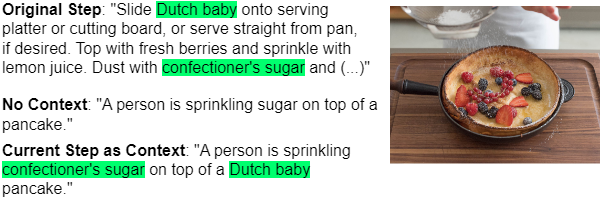}
    \caption{Example captions generated by InstructBLIP. In the "No Context" example, the model only receives the image. In the "Current Step as Context" example, the model receives the image plus the "Original Step".}
    \label{fig:caption-coreference}
\end{figure}

To train the decoder model $\varphi(\cdot)$, we generated contextual captions for each image in dataset $\mathcal{D}$ using InstructBLIP~\cite{InstructBLIP}.
To achieve richer and contextualized captions, we prompted InstructBLIP with additional context, conditioning the caption on the recipe steps, in addition to the image. Figure~\ref{fig:caption-coreference} shows example captions generated by the model, given a real data point in the dataset. 

\begin{figure*}[t]
  \centering
    \includegraphics[width=1.0\linewidth]{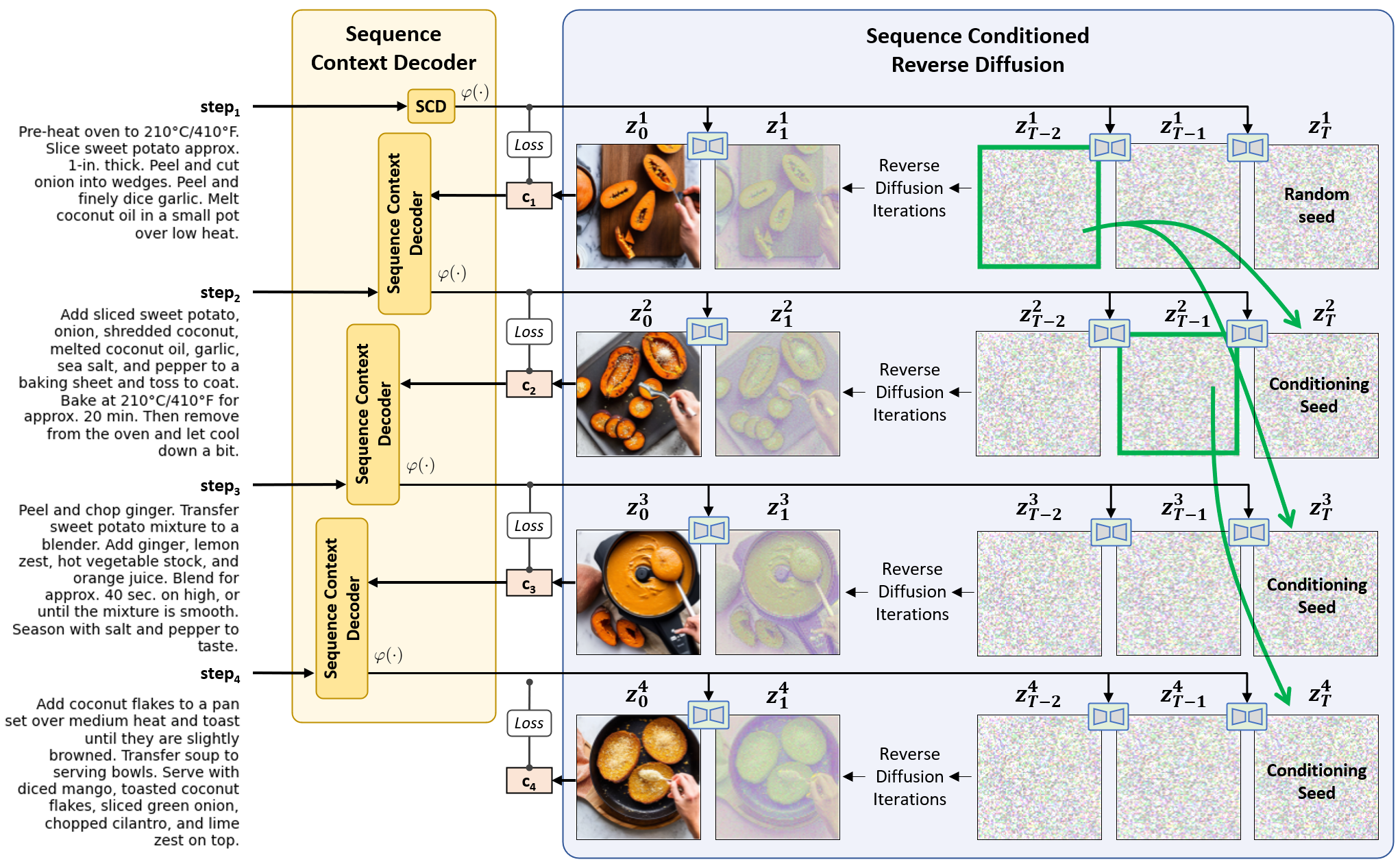}
  \caption{The proposed method uses the sequence context decoder to maintain semantic coherence. The reverse diffusion process uses a conditioning seed $z^{i}_T$ that is copied from a previous step and iteration $z^{j}_k$. See Equation~\ref{eq:loss_sldm}.}
  \label{fig:full-example}
\end{figure*}

\subsection{Sequence Conditioned Reverse Diffusion}
To maintain visual coherence among images in the sequence, we need to condition the current generation on the previous images. Image-to-image~\cite{Image2Image} generation follows this principle, but new images are too strongly influenced by previous ones and do not correctly integrate the new aspects present in the step description.

Following the rationale of conditioning every reverse diffusion process on previous processes, we propose to leverage latent vector iterations from early reverse diffusion processes. This leads us to the final formulation of the proposed method,
\begin{multline}
        \mathcal{L}_{SLDM}(s_i) = \mathbb{E}_{E(v_i), s_i,\epsilon, t} \Bigl[ \| \epsilon - \epsilon_\theta(z^{i}_t, t, \\
        \tau_\theta(c_i = \varphi(s_i, \{ s_{i-1}, \ldots, s_{i-w}\}))) \| ^2_2 \Bigr]
    \label{eq:loss_sldm}
\end{multline}
where for each $s_i$ a new reverse diffusion process starts with a conditioning seed $z^{i}_T$ copied from a previous step $s_{j}$ with $j<i$ and a latent vector iteration $k$ corresponding to the latent vector $z^{j}_k$, as illustrated in Figure~\ref{fig:full-example}. Next, we describe the details of how these two variables are determined.

\subsubsection{Random and Fixed Seeds}
To ground the generation, a straightforward method is to set a fixed seed for every step $i$ in the sequence, $z^i_T = c^{te}$. By fixing the initial seed, we aim to improve the coherence between generated images. The first two columns in Figure~\ref{fig:all-examples} demonstrate this approach.
We observed a greater homogeneity between generated images when using a fixed seed. Hence, all step illustrations share the same random seed, to achieve more coherent scenes and backgrounds but without the capacity to select the optimal starting seed.

\subsubsection{Conditioned Initialisation}
While using a fixed seed can improve the results, we argue that a better solution is achieved by using latent vectors from previous reverse diffusion processes. In particular, \textit{latent vectors that have already been semantically conditioned on past steps}. 
Figure~\ref{fig:full-example} shows how the latent vector representations $z^i_t$ evolve with increasing iterations, until they arrive at the final image $v_i=D(z^i_0)$ for step $s_i$.
These latent representations already contain meaningful information about the image~\cite{GuidedImageSynthesis}, which could be leveraged to improve the coherence of the following generations.
To achieve this, we need to carefully select which step to choose, to use as input seed for the next step image generation.

A step $s_i$ may be dependent on any previous step $j<i$, $\{s_{i-1}, \ldots, s_{1}\}$. To select the optimal initialization of the reverse diffusion process for step $s_i$, we start by determining the most similar step $s_j$ as the
\begin{equation}
    \argmax_{j} sim(s_i, s_j \in \{ s_{i-1}, \ldots s_{1} \})
\end{equation}
where $sim(\cdot)$ represents CLIP text similarity. If this similarity score is above a predefined threshold $\eta$, we use $s_j$ to extract the latent vector. If no step $s_j$ has a similarity score above $\eta$, we generated image $v_i$ with the shared random seed.

The reverse diffusion process progressively iterates over the latent vectors towards the final image. This means that conditioning the reverse diffusion process on latent vector iterations from a later iteration, i.e. a highly denoised latent vector, would force the resulting image to be very close to the previous one.
To decide how strongly we want to condition $v_i$ on the step $s_j$, we select the $k^{th}$ latent vector iteration as 
\begin{equation}
    k = n \cdot (sim(s_i, s_j) - \eta) / (1.0 - \eta)
\label{eq:similarity-interpolation}
\end{equation}
where $n$ is the maximum number of reverse diffusion iterations that we consider.

This brings us to the target reverse iteration vector $z^j_k$ which will be used as a starting seed $z^i_T$ in Eq~\ref{eq:loss_sldm} when calculating $\mathcal{L}_{SLDM}(s_i)$.
Figure~\ref{fig:full-example} illustrates the whole process: the proposed method captures the visual aspects that should be in the image, and the linked denoising latent vector provide the seed to generate a step image.

\section{Experimental Methodology}
\paragraph{Dataset.}
\label{sub:dataset}
We collected a dataset consisting of publicly available manual tasks in the recipes domain from \href{https://www.allrecipes.com/}{AllRecipes}. We also considered DIY manual tasks from \href{https://www.wikihow.com/}{WikiHow}, in an out-of-domain evaluation.
Each manual task has a title, a description, a list of ingredients/resources, and a sequence of step-by-step instructions, which may or may not be illustrated. 
Since we want to illustrate the steps of a task, we focus on manual tasks which are fully illustrated, as we can use these images as ground-truth for training and evaluating our methods. 
In total, we used 1100 tasks, with an average of 5.06 steps per task, resulting in 5562 individual steps.

\paragraph{Contextual Caption Generation.}
\label{sub:experimental-caption-generation}
As previously described, we provide InstructBLIP~\cite{InstructBLIP} with additional context to produce contextualized captions. We experimented with different context lengths and decided to rule out experiments that gave InstructBLIP the full context, i.e., all previous step instructions, as this led to very long outputs that often repeated irrelevant information from the input context, instead of describing the image.
We addressed the issue of long input prompts, by using a context window, as described in Section~\ref{subsub:task-context}. This allows us to give the model additional context while mitigating possible errors in the generated training caption.
When generating the image captions used to train the Sequence Context Decoder, we produced two sets of captions: long and short. The prompt to InstructBLIP consists of the additional context window followed by \textit{"Given the steps, give a short description of the image. Do NOT make assumptions, say only what you see in the image."}.
We generate captions with a window of 2 steps--short captions--and with a window of 3 steps--long captions.
Finally, we generate a long and a short caption for every image, $v_i$ in the dataset.

\paragraph{Model Details.}
To train the sequence context decoder model, we fine-tuned an Alpaca-7B model for 10 epochs on a single A100 40Gb GPU. We used a cross-entropy loss and a cosine learning rate scheduler, starting at 1$e^{-5}$. The batch size was set to 2, with a gradient accumulation step of 4. The dataset had a total of 5562 step-caption pairs, from which we used 80\% for training and 180 examples for testing. We used a frozen Stable Diffusion 2.1~\cite{stable-diffusion} for image generation.

\paragraph{Human Annotations.}
To evaluate our models we ran three annotation jobs, and, in all cases, annotators were allowed to provide feedback on the generation errors. See Appendix~\ref{sec:annotations} for details.
In the first job, annotators inspected 30 sequences of images generated with different methods (withheld from annotators) and selected the 3 best sequences, out of a total of 5 sequences. Besides this selection, we provided an additional \textit{No good sequence} label, for when no sequence of images was of good quality.
The second job aimed at obtaining finer-grained annotations for two methods that performed best.
Annotators compared the proposed method against the second best method (latent 1) and were asked to select the preferred one or to indicate a tie.
After these two initial human annotation jobs, we decided to compare the proposed method to the real-world image sequences and asked annotators to rate the two sequences. 
This is by far the most challenging setting where real-world image sequences have the natural visual coherence that we wish to achieve.

\begin{figure}[t]
    \centering
    \includegraphics[width=0.9\linewidth,trim={0 0 0 2pt},clip]{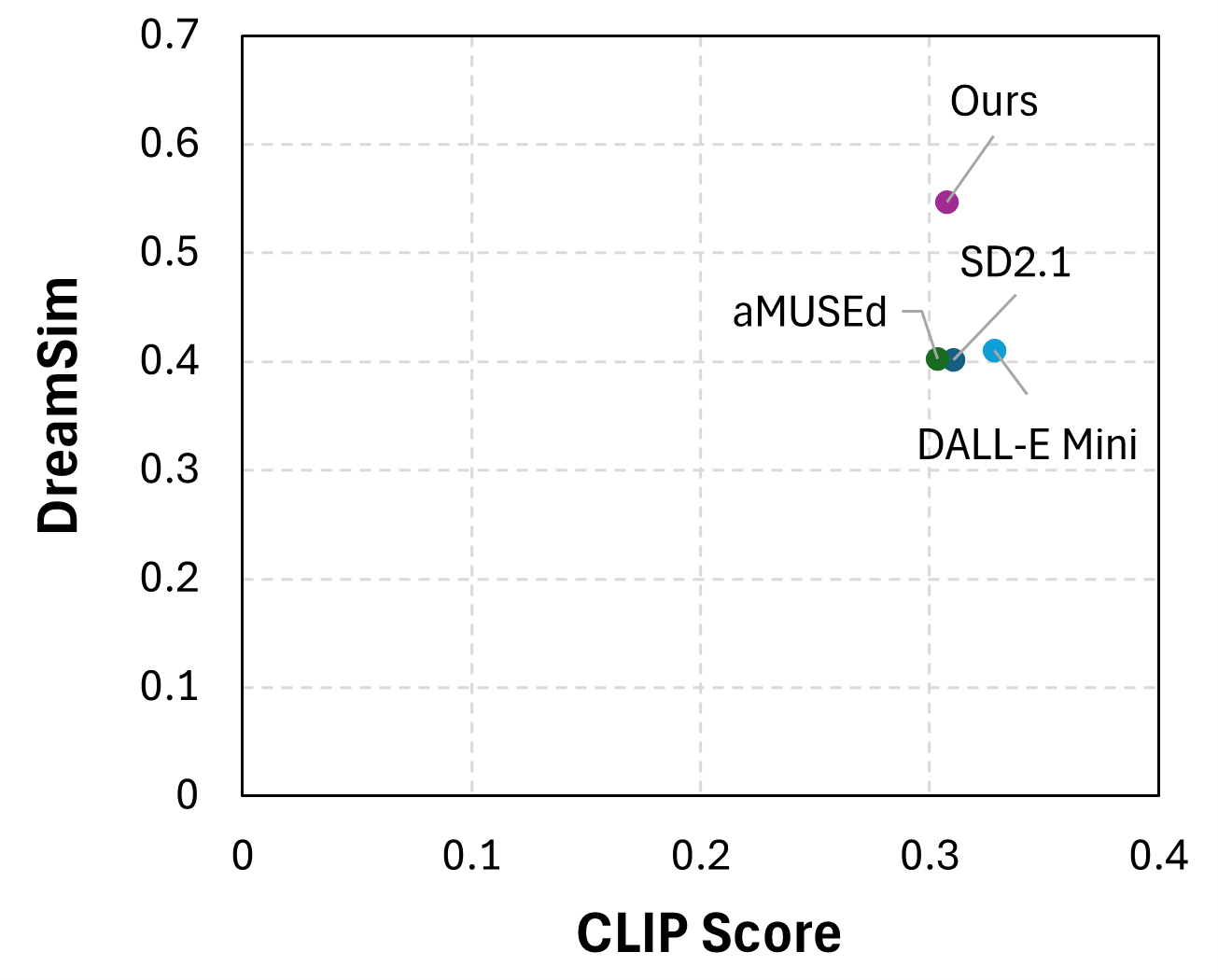}
    \caption{Automatic evaluation of image sequence. CLIP-Score~\cite{CLIPScore} measures the alignment between the step and the image. DreamSim~\cite{dreamsim} measures similarity between visual illustrations in the sequence.}
    \label{fig:autoeval}
\end{figure}

\paragraph{Automatic Metrics.}
To assess image sequences generated by the proposed method, we measured the semantic correctness and the sequence coherence with CLIPScore~\cite{CLIPScore} and with the novel DreamSim metric~\cite{dreamsim}, respectively. DreamSim measures the similarity between two images in terms of their foreground objects and semantic content, while also being sensitive to colour and layout. This is particularly well suited to our task, as we wish to maintain visual coherence across the entire sequence.

\section{Results and Discussion}
In this section, we start by comparing the proposed method to latent diffusion models, SD2.1~\cite{stable-diffusion}, and non-LDM models, i.e. aMUSEd~\cite{amused} and DALL·E Mini
~\cite{Dayma_DALL·E_Mini_2021}, with automatic metrics.
For the ablation studies, we experimented with using (1) a \textbf{random seed} for all steps of the sequence, (2) a \textbf{fixed seed} for all steps of the sequence, (3) a \textbf{fixed latent vector iteration} from the previous step, represented by \textbf{Latent $k$}, where $k$ is the fixed iteration, and (4) the proposed method that selects a latent vector iteration, $z^j_i$ from the previous denoising steps as a starting seed.

\subsection{Automatic Evaluation}
\label{sub:autoeval}
Figure~\ref{fig:autoeval} shows that the proposed method can improve the coherence of image sequences (as measured by DreamSim), while maintaining the same text-to-image generation capabilities (as measured by CLIPScore).
This result is particularly important because it clearly shows that it is possible to maintain key visual and semantic traits from specific iterations of the reverse-diffusion process -- an LDM property that we leverage in this paper.

\begin{table}[t]
\centering
\begin{tabular}{@{}lrrr@{}}
\toprule
\textbf{Method} & \textbf{Best (\%)} & \begin{tabular}[r]{@{}r@{}}\textbf{Second}\\ \textbf{Best (\%)}\end{tabular} & \begin{tabular}[r]{@{}r@{}}\textbf{Third} \\ \textbf{Best (\%)}\end{tabular} \\ \midrule
Random seed & 17.70 & \textbf{41.20} & 13.00 \\
Fixed seed & \underline{29.40} & 17.70 & \underline{33.30} \\
Latent T-1 & \textbf{33.30} & 17.70 & \textbf{37.04} \\
Latent T-2 & 17.70 & \underline{23.50} & 16.70 \\
Img-to-Img & 2.00 & 0.00 & 0.00 \\ \bottomrule
\end{tabular}%
\caption{Annotation results for the evaluation of the various methods of maintaining visual coherence. Annotators picked \textit{No Good Sequence} in 18.99\% of the sequences; we report the results for the remaining 81.01\%.\\}
\label{tab:full-results}
\begin{tabular}{lcc}
\toprule
\multirow{2}{*}{\textbf{Method}} & \textbf{Recipes}  & \textbf{DIY}\\ 
                                 & \small{(seen)}  & \small{(unseen)}\\ 
\midrule
Proposed method \small{(wins)} & \textbf{46.67} & \textbf{30.00} \\
Second best \small{(wins)}  & 26.67 & 23.33 \\
Tie & 10.00 & 16.67 \\
No good sequence & 16.67 & 30.00 \\
\bottomrule
\end{tabular}%
\caption{Annotation results of the comparison between our proposed method and the winning method from Table~\ref{tab:full-results} (Latent T-1).}
\label{tab:heuristic-vs-latent}
\end{table}

\begin{figure*}[!ht]
    \centering
    \begin{minipage}{\textwidth}
        \centering
        \includegraphics[width=0.95\linewidth]{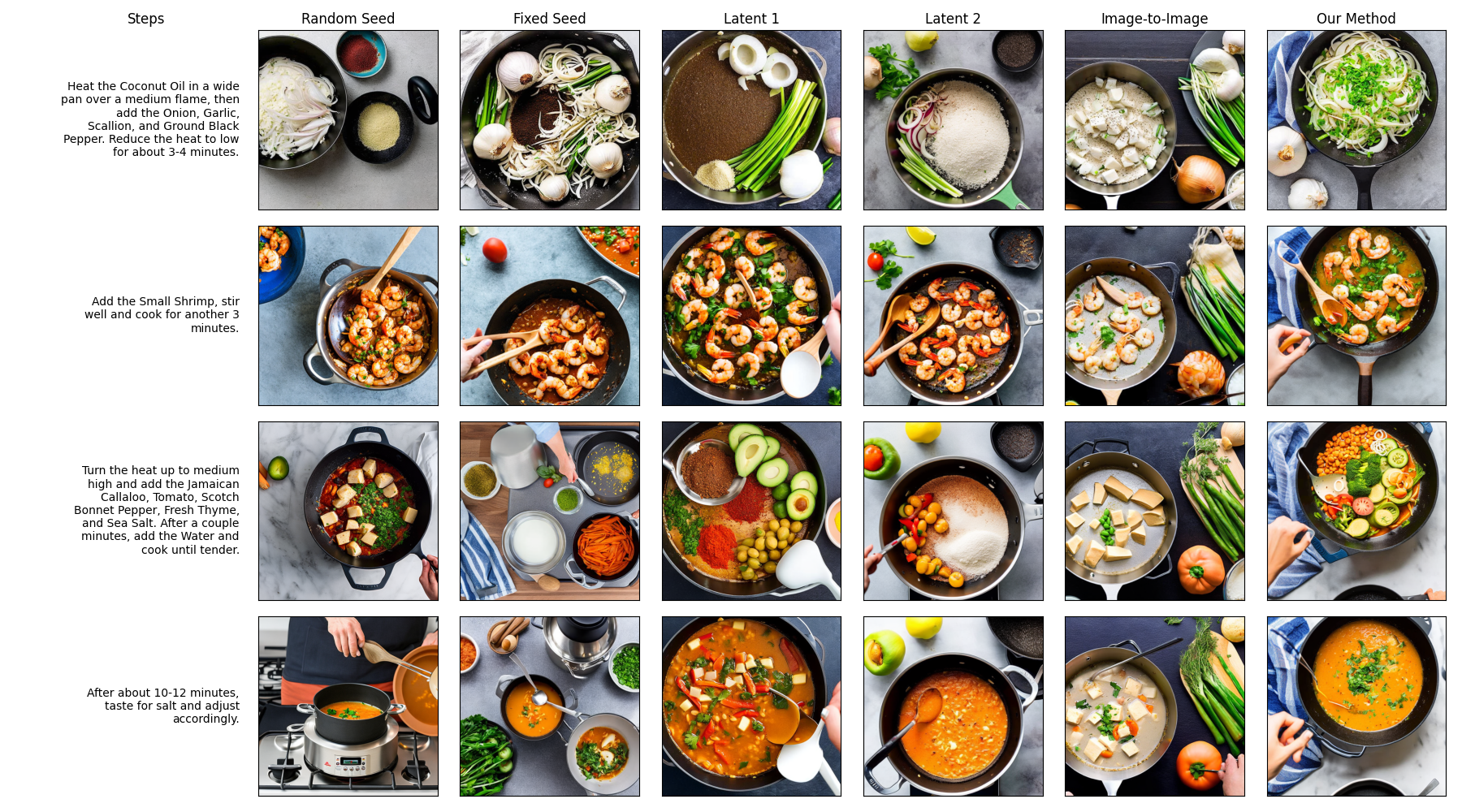}
        \caption{Examples of recipe illustrations with different methods for maintaining visual coherence.}
        \vspace{-2mm}
        \label{fig:all-examples}
    \end{minipage}
\end{figure*}
\subsection{Sequence Generation Results}
\label{sub:baselines}
\textbf{Recipes Domain.} To validate our initial hypothesis, we start by analysing its performance in the recipes domain. Our goal is to assess how conditioning the reverse diffusion process of each step affects the overall results. 
Table~\ref{tab:full-results} provides the complete set of results across all competing methods. It is clear how using latent vector iterations and random seeds supports the intuition behind our method: a manual task is composed of continuous and independent actions, which should be conditioned by latent vector iterations and by random seeds, respectively.
Building on this observation, we calibrated the $\eta$ parameter of the proposed method, using human evaluation, as reported in annex in Table~\ref{tab:heuristic-tuning}.

To compare the proposed method to the best performing method (Latent T-1), we asked annotators to select the best sequence out of two side-by-side sequences (see Appendix~\ref{sec:annotations} for details about the annotation instructions). 
Results reported in Table~\ref{tab:heuristic-vs-latent} show that in $46.7\%$ of the cases, human annotators preferred our method over the alternative and in $10.0\%$ of the cases the two methods were equally good. 
According to the annotations, the proposed method was equal or better than the second best baseline with an agreement of 70\% across the full set of tasks and annotators.
This confirms our hypothesis and supports the importance of selectively conditioning the denoising process on the previously generated steps of the sequence.

\paragraph{DIY Domain.}
To assess the generalisation of the proposed method, we evaluated its performance in an unseen domain: DIY tasks.
With the results of our human annotation study, we observed that the transition from generating recipe images to DIY tasks has shown promising results.
Results in Table~\ref{tab:heuristic-vs-latent} show that in 30.0\% of the tasks, neither method produced satisfactory results. For the tasks that were correctly illustrated, we see that annotators preferred our method in 30.0\% of the tasks, compared to 23.3\% for the second-best approach. Additionally, 16.7\% of comparisons resulted in a tie between the two methods. Although we see limitations in this domain, the results show that our approach is capable of generalizing to an unseen domain, producing satisfactory image sequences.

\paragraph{Generated vs Ground-Truth Sequences.}
We compared the quality of generated image sequences against the ground-truth sequences and asked human annotators to rate each sequence in a 5 point Likert scale. This is a particularly challenging setting because the real image sequences are photos taken by humans in a real-world setting, where sequence coherence is naturally captured. Table~\ref{tab:our-method-vs-ground-truth} shows that our method achieves over $60\%$ of the ground-truth score, with the ground-truth sequences only $0.42$ points below the maximum score.
\begin{table}[t]
\centering
\begin{tabular}{@{}ll@{}}
\toprule
\textbf{Method}       & \textbf{Average Rating}       \\ \midrule
Proposed method   & 2.93 $\pm$ 1.14 \\
Ground-truth & 4.58 $\pm$ 0.79 \\ \bottomrule
\end{tabular}
\caption{Human annotation results for the comparison of the proposed method with ground-truth images.}
\label{tab:our-method-vs-ground-truth}
\vspace{-2mm}
\end{table}

\paragraph{Qualitative Analysis.}
To illustrate how different conditioning methods affect the quality of generated sequences, we present several examples in Figure~\ref{fig:all-examples} and in Appendix~\ref{sec:examples_analysis}.
In Figure~\ref{fig:all-examples}, we can see that by using a fixed seed for all steps, we are able to preserve the background and add objects for each specific step. We can also see that the image-to-image method conditions the generations too strongly.
Figure~\ref{fig:all-examples} also shows that our method is capable of preserving and recall key visual artefacts from several steps back in the sequence. We believe this is a distinctive and fundamental feature of the proposed method. 

In the DIY out-of-domain experiment, Figure~\ref{fig:diy-examples} in annex shows a strong generalization to tasks involving simple object utilisation, such as using a broom for cleaning or a brush for painting. While fixed latent vector iteration methods show some memorization capability, the image-to-image generations are too biased on previous generations, and random seeds lead to very diverse generations. 
In this unseen domain, the proposed method encounters considerable challenges when tasked with more intricate activities, like performing a car's oil change with its complex mechanical components, or engaging in tasks that involve philosophical or introspective elements, see Appendix~\ref{sec:examples_analysis} for visual examples. It is worth noting that these limitations are inherited from the core image generation method, which struggles to handle fine-details.

\subsection{Sequence Conditioned Reverse Diffusion}
\label{sub:seq_cond_rev_dif}
To better understand the strength of our \textbf{visual coherence} hypothesis, we conducted an experiment where all the images of the sequence are generated from the latent vectors of a fixed iteration from the previous task step. 
Specifically, we use the latent vector iteration $k$, with $k \in \{1,2,5,10,20,49\}$, of the previous task step as the starting point of the current reverse diffusion process.

Our empirical analysis of these generations, from which some examples can be seen in annex in Figure~\ref{fig:latents-various-steps}, confirms our initial hypothesis showing that using latent vector iterations from previous steps provides a good result in many cases.
Additionally, this experiment evidences the strength of latent vector iterations as conditioning signals in a reverse diffusion process.
This experiment shows that later iterations add a very strong bias to the generation emphasising the importance of selecting the best conditioning seed from previous reverse diffusion iterations and processes.
An extreme case of this phenomenon is observed in the image-to-image method in Figure~\ref{fig:all-examples}. 

\subsection{Sequence Context Decoder}
We assessed the capacity of the sequence context decoder of maintaining the \textbf{semantic coherence} by manually annotating the decoder output for six settings with different context lengths and caption lengths, Table~\ref{tab:sequence-context-decoder-results}. We considered three context lengths: the shortest one considers the current step, while the longest, shows two steps and a caption. For the captions, we used both \textit{short} and \textit{long} captions, as detailed in Section~\ref{sub:experimental-caption-generation}.

Table~\ref{tab:sequence-context-decoder-results} shows the results for the different evaluation settings. These results indicate that the model attains the best semantic coherence with a context window of 2 and with short captions. We observed that captions generated with short contexts tend to lack some information, while captions generated with longer contexts introduced too much information, which was often noisy. This is aligned with a recent study~\cite{liu2023lost} that highlights the fact that current large language models do not robustly make use of information in long input contexts.
These results indicate that additional context needs to be carefully considered and curated, as the model is not able to filter out excess information. In annex~\ref{sec:decoder_failure_analysis} we provide more insight into the performance of the decoder.

\begin{table}[t]
\centering
\begin{tabular}{@{}lC{2cm}C{1.2cm}@{}}
\toprule
\textbf{Sequence Context} & \textbf{Captions} &  \textbf{Avg. Rating}\\
\midrule
\{s$_n$\} & \textit{short} & $3.00$\\
\{s$_n$\} & \textit{long} & $3.23$\\
\midrule
\{s$_n$, c$_{n-1}$\} & \textit{short} & $\textbf{3.68}$\\
\{s$_n$, c$_{n-1}$\} & \textit{long} & \underline{$3.56$}\\
\midrule
\{s$_n$, s$_{n-1}$, c$_{n-2}$\} & \textit{short} & $3.41$\\
\{s$_n$, s$_{n-1}$, c$_{n-2}$\} & \textit{long} & $3.35$\\ \bottomrule
\end{tabular}
\caption{Sequence Context Decoder results for different context lengths, configurations and caption type.}
\vspace{-2mm}
\label{tab:sequence-context-decoder-results}
\end{table}

\section{Conclusions}
In this paper, we addressed the problem of illustrating complex manual tasks and proposed a framework for generating a sequence of images that illustrate the manual task. The framework is composed of a novel \textbf{sequence context decoder} that preserves the \textbf{semantic coherence} of a sequence of actions by transforming it into a visual caption.
The full sequence illustration framework is completed by a \textbf{sequence conditioned reverse diffusion} process that uses a latent vector iteration from a past image generation process to maintain \textbf{visual coherence}.

Automatic and human evaluations in the target domain demonstrated the strong performance of the framework in generating coherent sequences of visual instructions of manual tasks. The proposed method was preferred by humans in 46.6\% of the cases against 26.6\% for the second best method. In addition, automatic measures, also confirmed that our method maintains visual and semantic coherence while illustrating a complex manual task.
The generalization to unseen domains was successfully validated in the DIY domain with positive results.

\section{Risks and Limitations}
While we conducted a thorough set of experiments and validations, we acknowledge that more experiments could shed more light in some aspects. First, we did not experiment with larger LLMs to discover the impact of scaling. Second, we also did not consider that steps may dependent on multiple previous steps. Third, the proposed solution only considers a limited window of steps as context, which may lead to some information being lost. 

Finally, in terms of risks, we acknowledge that our work could potentially be used to generate false information.

\bibliography{bibliography}
\bibliographystyle{acl_natbib}

\clearpage
\appendix

\begin{figure*}[!ht]
    \vspace{4mm}
    \begin{minipage}{\textwidth}
        \centering
        \includegraphics[width=\linewidth]{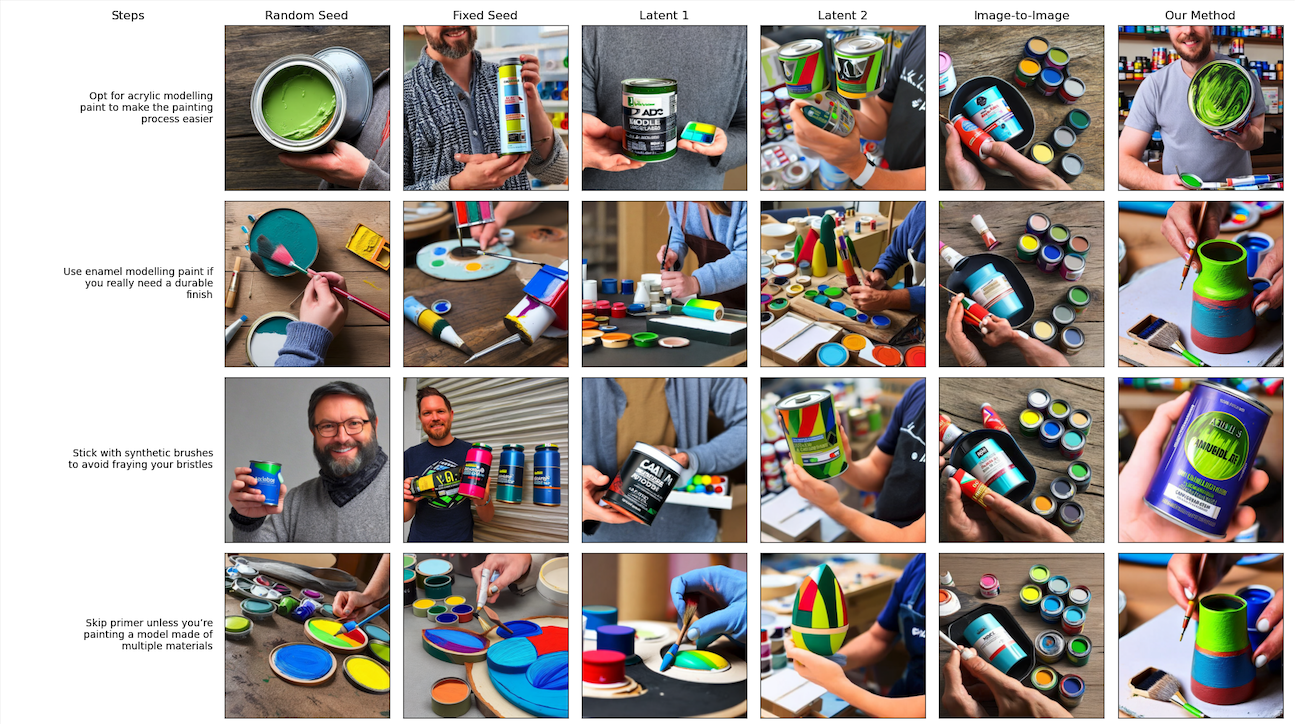}
        \caption{Examples of DIY illustrations with different methods for maintaining visual coherence. Each column illustrates steps 1 through 4.}
        \label{fig:diy-examples}
    \end{minipage}
    \\
\end{figure*}

\section{Data Preparation}
\label{sec:appendix}
We found that recipes with long steps descriptions or many steps, were difficult to tackle and produced worse results. This pointed towards a refinement of the dataset, so that our approach could better focus on the issue of coherence, instead of tackling other problems.

When a step description is too long, it contains too much information, often with multiple actions, which is hard to represent in a single image. Adding to this issue, the CLIP~\cite{CLIP} text encoder, used in the Stable Diffusion~\cite{stable-diffusion} model, truncates the input text at 77 tokens.
We filtered the recipes that had steps that were too long.
A second problem arises when a recipe has too many steps, as it is difficult to produce coherent illustrations over such a long sequence of steps. To mitigate this concern, we limited the number of steps in a recipe from 4 to 6 steps. In a final stage of refining the dataset, we removed any steps that did not contain actions which we could illustrate, such as steps merely saying \textit{“Enjoy!”}.

To the best of our knowledge, we do not use any personal identifiable information.

We plan to release the data under an Apache-2.0 licence. We intend to allow future work to further research the problem, using the provided data.

\section{Model training}

\begin{table}[]
\centering
\begin{tabular}{@{}ll@{}}
\toprule
\multicolumn{2}{c}{\textbf{Training Details}} \\ \midrule
Base Model                   & Alpaca-7B      \\ \midrule
Training Time                & $\approx 10h$    \\
Epochs                       & 10             \\
Loss Function                & Cross-Entropy  \\
Weight Decay                 & 0.01           \\
Model Max Length             & 400            \\ \midrule
Batch Size                   & 2              \\
Gradient Accumulation Steps  & 4              \\
Effective Batch Size         & 8              \\ \midrule
Learning Rate                & 1$e^{-05}$          \\
Learning Rate Scheduler      & Cosine         \\ \midrule
Optimizer                    & AdamW          \\ 
\hspace{3mm}Adam $\beta_1$     & 0.900          \\
\hspace{3mm}Adam $\beta_2$     & 0.999          \\
\hspace{3mm}Adam $\epsilon$    & 1$e^{-08}$           \\ \midrule
LoRA                         &  {}          \\
\hspace{3mm}LoRA Rank        & 8             \\
\hspace{3mm}LoRA $\alpha$      & 32             \\
\hspace{3mm}LoRA Dropout     & 0.1            \\
\bottomrule
\end{tabular}
\caption{Training parameters for the best model.}
\label{tab:training-parameters}
\end{table}

We chose Alpaca-7B as our base model, as it is an open-source instruction-tuned model, and there are implementations using LoRA~\cite{LoRA}, which reduces the computational cost during training. We encourage future work to study the impacts of scaling the LLM. We experimented with different hyperparameters, namely different learning rates, and learning rate schedulers, to find the most suitable ones for our problem. We point out that we are using the loss of the models as the main criterion for future experiments, as we do not have an automatic metric for evaluating model behaviour. The loss is not always indicative of the model's performance in the task at hand, as we verified empirically.

\begin{table}[ht]
\centering
\begin{tabular}{lrrr}
\toprule
$\eta$ & \textbf{Best (\%)} & \begin{tabular}[c]{@{}r@{}}\textbf{Second} \\ \textbf{Best (\%)}\end{tabular} & \begin{tabular}[c]{@{}r@{}}\textbf{Third} \\ \textbf{Best (\%)}\end{tabular} \\ \midrule
0.70 & 19.20 & 12.80 & 14.90 \\
0.65 & 12.80 & 14.90 & \textbf{25.50} \\
0.60 & 19.20 & 23.40 & 21.30 \\
0.55 & 14.90 & 23.40 & 21.30 \\
0.50 & \textbf{34.04} & \textbf{25.53} & 17.02 \\ 
\bottomrule
\end{tabular}%
\caption{Annotation results for the sequences generated with different threshold values, $\eta$. Annotators picked \textit{No Good Sequence} in 20.34\% of the generations; we show results for the remaining 79.66\%.}
\label{tab:heuristic-tuning}
\end{table}

Since the cosine scheduler has greater variability between runs, due to different number of epochs leading to different loss curves, we decided to run further experiments with the next best-performing model. We experimented with varying the weight decay parameter but found that there were no significant differences in the loss curves for the three weight decay values.

For the aforementioned runs, we used $\beta_1 = 0.900$ and $\beta_2 = 0.999$ as the AdamW optimizer's hyperparameters. As a final test, we changed these to the values proposed by \citet{InstructGPT}, $\beta_1 = 0.900$ and $\beta_2 = 0.950$. We did not see any improvement in the loss curve.

Based on these results, we fine-tuned our Alpaca-7B models for 10 epochs on a single A100 40Gb GPU. We used a cross-entropy loss, a cosine learning rate scheduler, starting at 1$e^{-5}$. Our batch size was 2, with a gradient accumulation step of 4, leading to an effective batch size of 8. The dataset had a total of 5562 examples; we used 80\% for training and the remaining for evaluation. Figure~\ref{tab:training-parameters} summarizes the training information for our best-performing model.

\section{Fixed Latent Iteration Generation}

\begin{figure*}[t]
    \includegraphics[width=\linewidth]{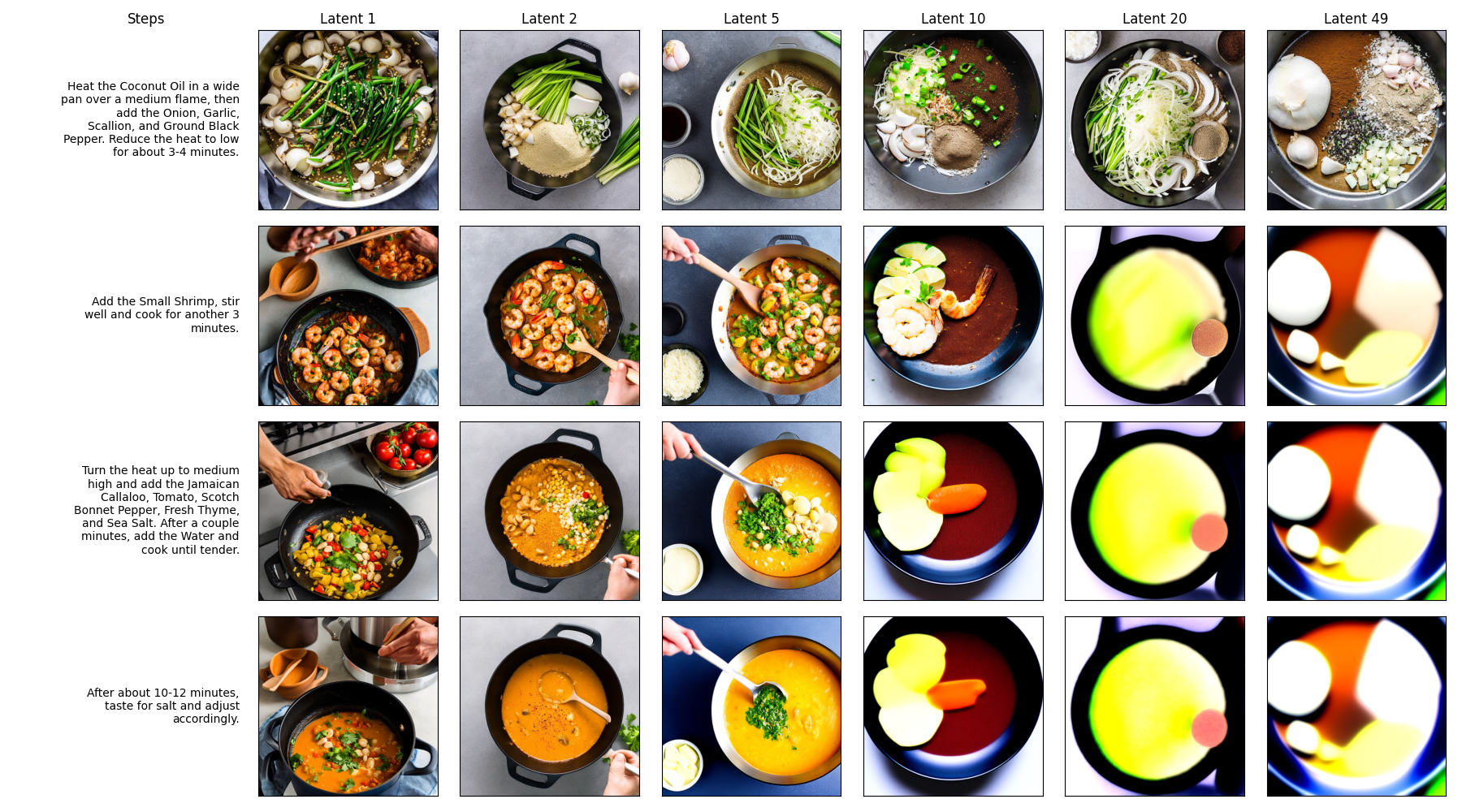}
    \caption{Maintaining visual coherence through the use of different memory latent vectors.}
    \centering
    \label{fig:latents-various-steps}
\end{figure*}

As mentioned in Section~\ref{sub:seq_cond_rev_dif}, we conducted an empirical analysis of generations using a fixed latent vector iteration from the previous step. This analysis helped us understand how different latent vector iterations impact the generation of the following image. An example from this analysis is shown in Figure~\ref{fig:latents-various-steps}.

\section{Negative Prompts}
We found problems which were not related to the prompts or the concepts we were trying to generate, but general problems in image generation, i.e., tiled images, or deformed hands. In order to reduce some of these common problems, present in Stable Diffusion generations, we used negative prompts. A negative prompt steers the generation away from the concepts present in it. This string of text is added to the end of the original prompt.

In the negative prompt, we included undesirable concepts such as \textit{human} or \textit{hands}, and also included some additional concepts, following common practices. Our final negative prompt follows: \textit{negative\_prompts = ["hands", "human", "person", "cropped", "deformed", "cut off", "malformed", "out of frame", "split image", "tiling", "watermark", "text"]}.

\section{Human Annotations}
\label{sec:annotations}

In this Section, we present examples of the annotation tasks.

The human annotation pool consisted of 3 PhD students and 5 MSc students. 25\% of the annotators were women and 85\% were men.

\begin{figure}[t]
  \centering
    \includegraphics[width=\linewidth, fbox]{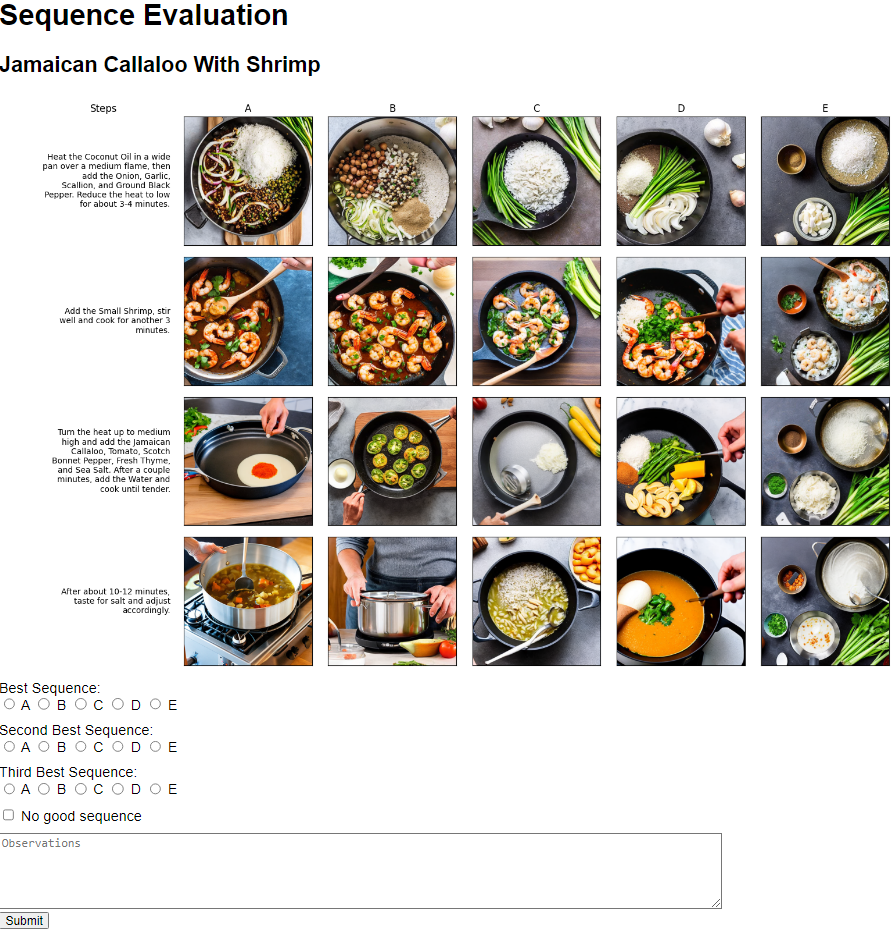}
  \caption{Annotation of the comparison between visual coherence methods.}
  \label{fig:visual-coherence-annotations}
\end{figure}

\begin{figure}[t]
  \centering
    \includegraphics[width=\linewidth, fbox]{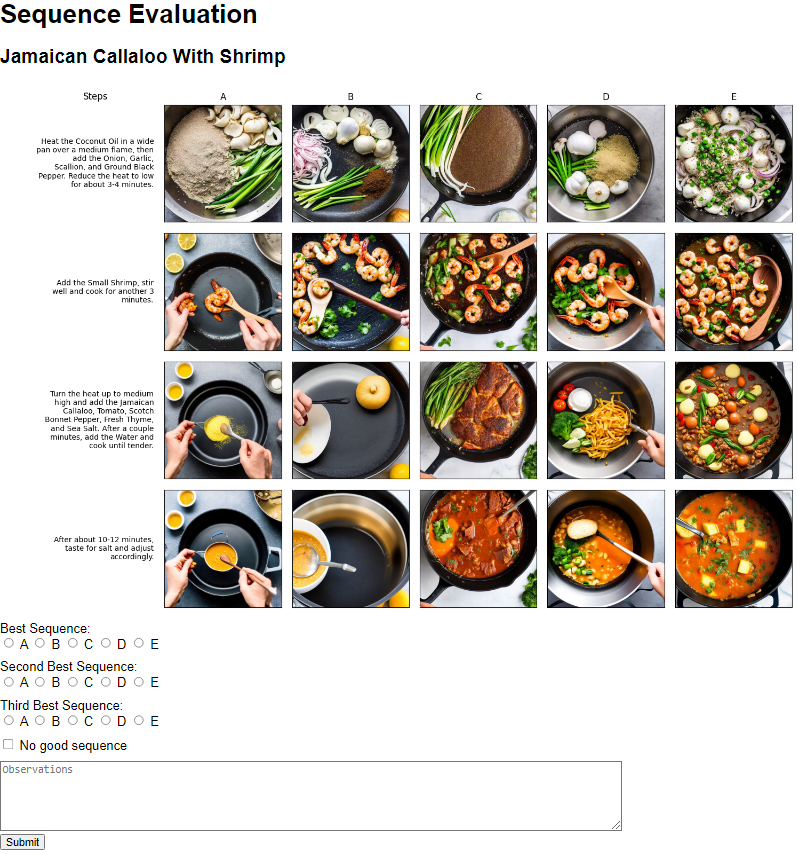}
  \caption{Annotation of the comparison between different heuristic thresholds.}
  \label{fig:heuristic-tuning-annotations}
\end{figure}

\begin{figure}[t]
  \centering
    \includegraphics[width=\linewidth, fbox]{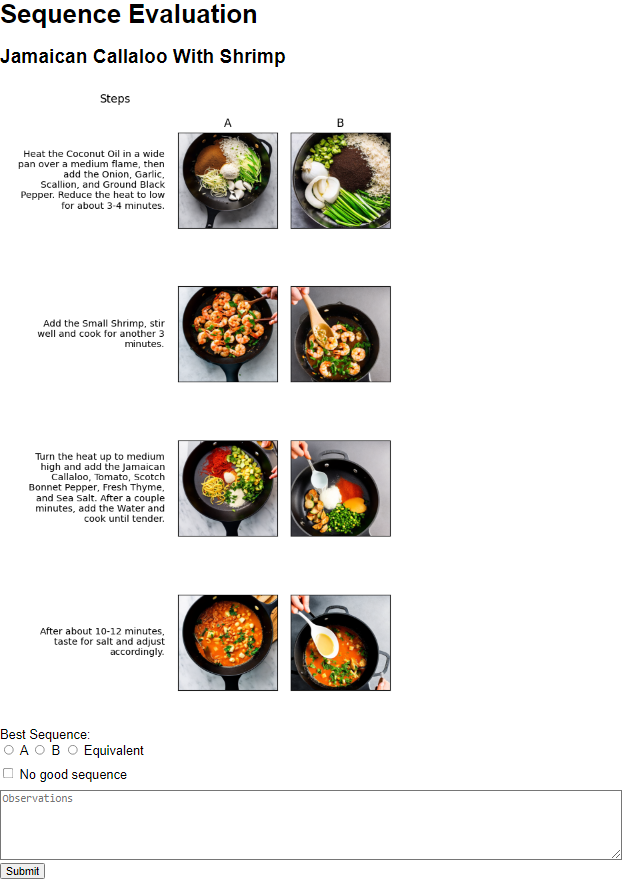}
  \caption{Annotation of the comparison between our method and the best visual coherence method.}
  \label{fig:heuristic-vs-latent-annotations}
\end{figure}

\begin{figure}[t]
  \centering
    \includegraphics[width=\linewidth, fbox]{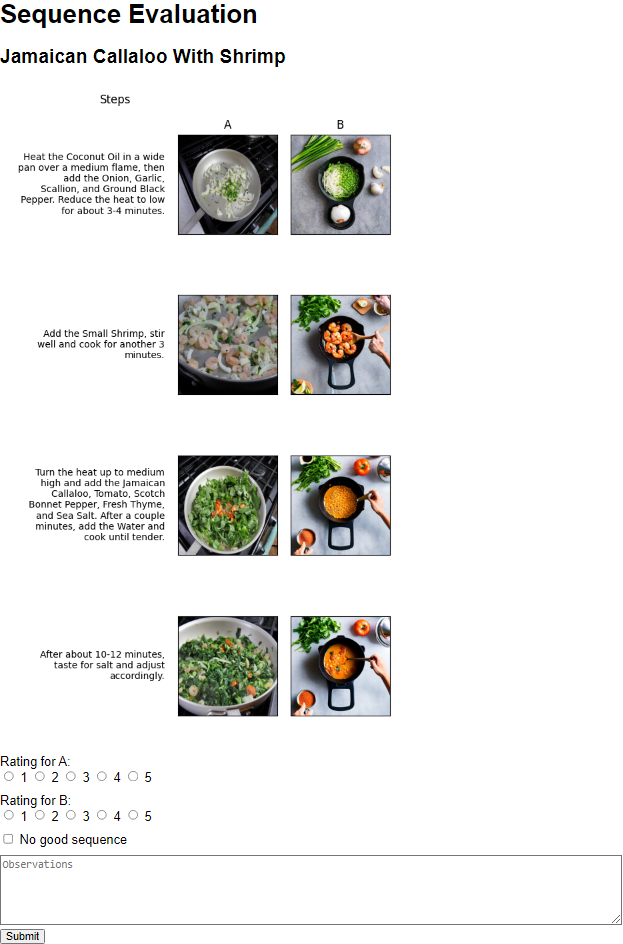}
  \caption{Annotation of the comparison between our method and the ground-truth images.}
  \label{fig:our-method-vs-ground-truth-annotations}
\end{figure}

\begin{figure}[t]
  \centering
    \includegraphics[width=\linewidth, fbox]{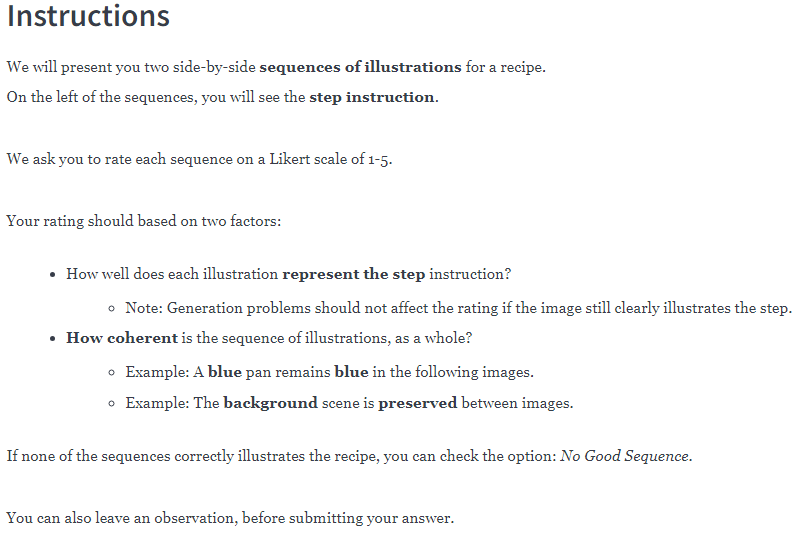}
  \caption{Annotation guidelines for the comparison between our method and the ground-truth images.}
  \label{fig:our-method-vs-ground-truth-annotation-guidelines}
\end{figure}

Figure~\ref{fig:visual-coherence-annotations} shows the annotation task to choose the best visual coherence maintaining method. The annotators saw 5 sequences: \textit{Random Seed}, \textit{Fixed Seed}, \textit{Latent 1}, \textit{Latent 2}, and \textit{Image-to-Image}. They were then asked to pick the best, second best, and third best sequences. They could also indicate that there were no good sequences, by checking the \textit{No good sequence} checkbox. Additionally, they could leave an observation, in the appropriate text area.
Figure~\ref{fig:heuristic-tuning-annotations} shows the annotation task to tune the threshold of our method. The annotators had to pick between 5 sequences, generated with different values of threshold: $0.50$, $0.55$, $0.60$, $0.65$, $0.70$. They were asked to pick the best, second best, and third best sequences. They could also indicate that there were no good sequences, by checking the \textit{No good sequence} checkbox. They could also leave an observation, in the appropriate text area.
Figure~\ref{fig:heuristic-vs-latent-annotations} shows the annotation task to choose between our method and the winning visual coherence maintaining method. The annotators saw 2 sequences, one generated with \textit{Latent 1} and another with our method. They had to choose the win sequence, or deem them equivalent. If there was no good sequence, they could check the \textit{No good sequence} checkbox. They could also leave an observation. Figure~\ref{fig:our-method-vs-ground-truth-annotations} shows the annotation task to rate sequences generated with our method and the ground-truth images, from a scale of 1 to 5. Additionally, the annotators could select that there was no good sequence, or leave an observation. Figure~\ref{fig:our-method-vs-ground-truth-annotation-guidelines} shows the annotation guidelines for the task to rate sequences generated with our method and the ground-truth images.

\section{Sequence Context Decoder Failure Analysis}
\label{sec:decoder_failure_analysis}
We further analysed the errors of the best method and present the results in Table~\ref{tab:error-types}. Hallucinations occurred in 3.9\% of the generations, and the LLM copied the input into the output in 7.2\% of the cases. 
Finally, it is interesting to see that the input was too complex in 6.2\% of the cases, i.e., describing more actions than what is possible to depict in the image.
\begin{table}[h]
    \centering
    \begin{tabular}{L{5.7cm}R{1cm}}
        \toprule
        \textbf{Error type}	& \textbf{\%} \\
        \midrule
        Hallucinations & 3.9\% \\
        Complex step with many actions & 6.2\% \\
        Copied input & 7.2\% \\
        \bottomrule
    \end{tabular}
    \caption{The contribution of each error type to the overall performance.}%, where in the first two errors are related to LDM's inherent hallucinations and to the complexity of the input sequence}
    \label{tab:error-types}
\end{table}

\section{Examples and Qualitative Analysis}
\label{sec:examples_analysis}

Table~\ref{tab:llm-generation-examples} shows some example generations from the Sequence Context Decoder, each highlighting a particular behaviour of the model. In Example 1, we can see the model correctly identifying the ingredients from the context, caption$_{n-2}$, going two steps back, and integrating them in the final output. It also recognizes the plate from step$_n$ as the object containing the ingredients. This shows the potential in giving the model additional context to generate richer prompts. In Example 2, we can see that, despite being able to maintain the \textit{red apples}, the model makes no explicit reference to their state, chopped up. This is still a limitation, which may lead to a wrongful representation of intact apples. In Example 3, we want to highlight two main aspects of the generation: we can see the model adding the \textit{bowl of soup} from the context to the prompt, maintaining semantic coherence. We can also see that the model kept \textit{lime juice}. This is correct, from the point of view of the task at hand, but shows the lack of understanding of what can be perceived in an image. We reason that this knowledge should come from the pretraining of the model, and not from our fine-tuning to this task. Example 4 shows an example of a depiction that is mostly correct, but misses a step of the sequence. The representation of the saucepan with black tea in it is plausible, but step$_n$ indicates the saucepan should be removed from the heat. Finally, in Example 5, we see a very long step, with various actions. In this case, we consider it plausible for the model to pick one of these actions. This is a better result than attempting to represent them all, which would lead to an inadequate prompt. Despite this, this specific generation lacks some context, as the word \textit{vegetables} is generic; it is important for the generated prompts to be specific, containing the ingredients mentioned in the context.

\begin{figure*}[t]
  \centering
    \includegraphics[width=1.0\linewidth]{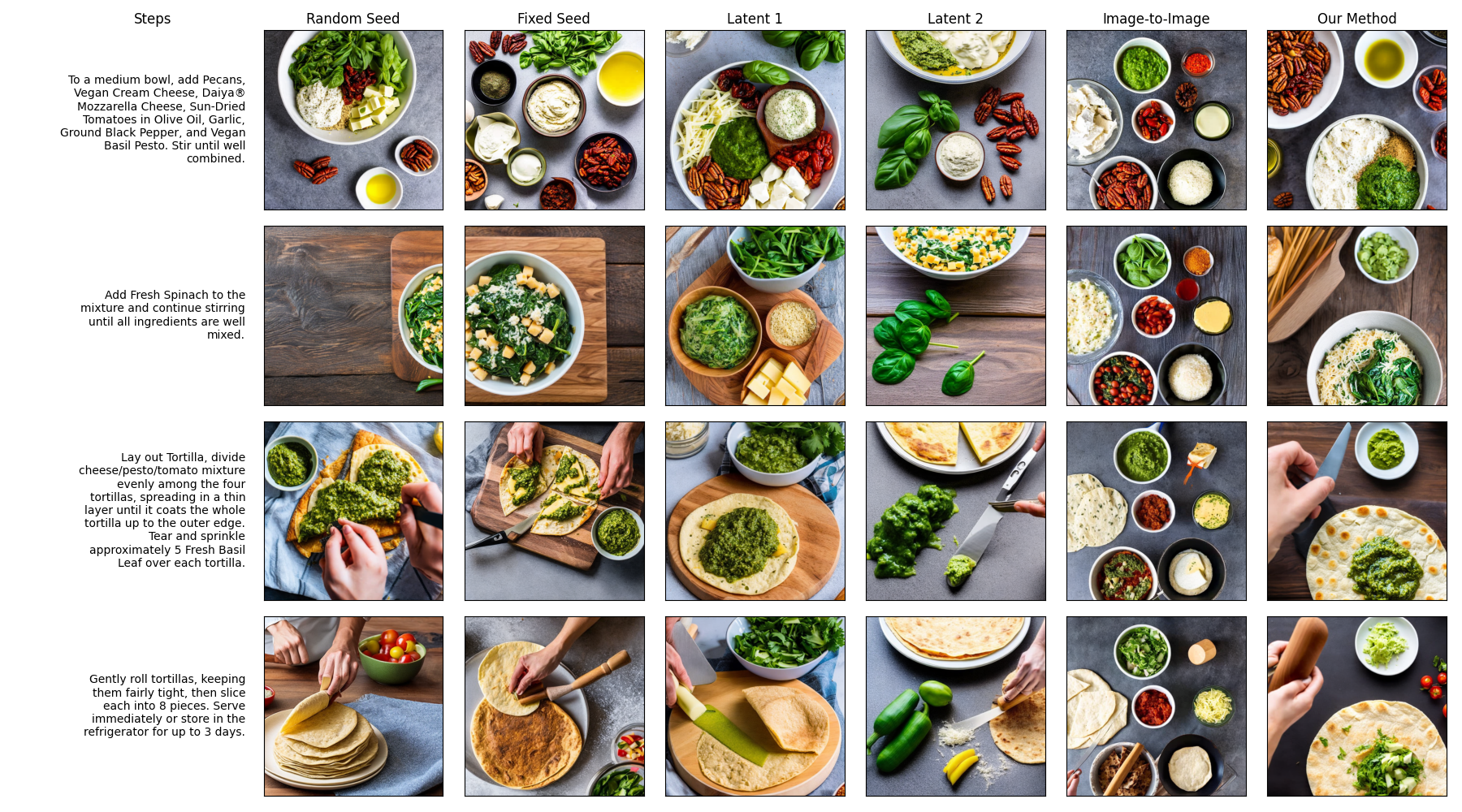}
  \caption{Examples of recipe illustrations with different methods for maintaining visual coherence.}
  \label{fig:visual-coherence-recipe-2}
\end{figure*}

\begin{figure*}[t]
  \centering
    \includegraphics[width=1.0\linewidth]{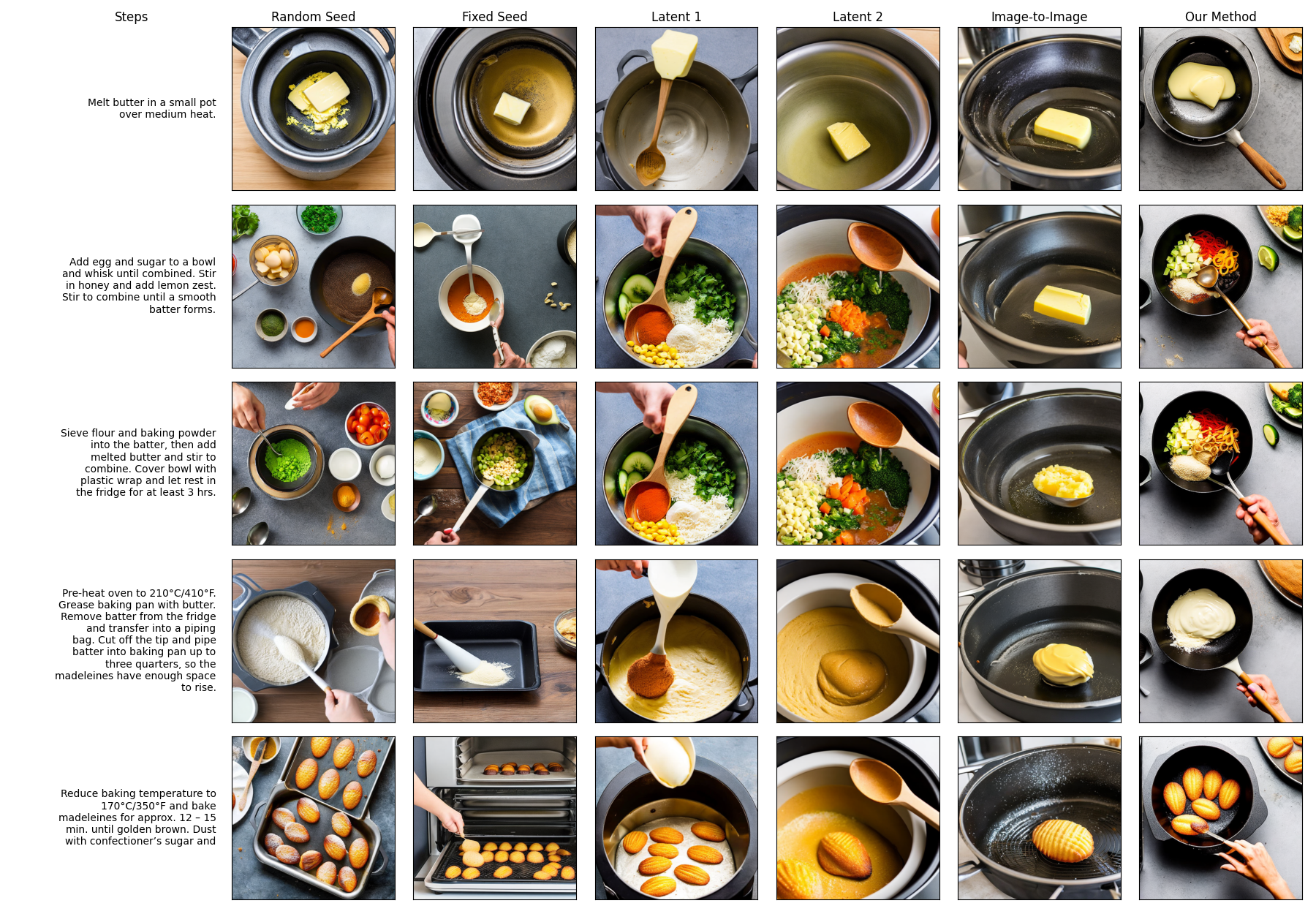}
  \caption{Examples of recipe illustrations with different methods for maintaining visual coherence.}
  \label{fig:visual-coherence-recipe-3}
\end{figure*}

\begin{figure*}[t]
  \centering
    \includegraphics[width=1.0\linewidth]{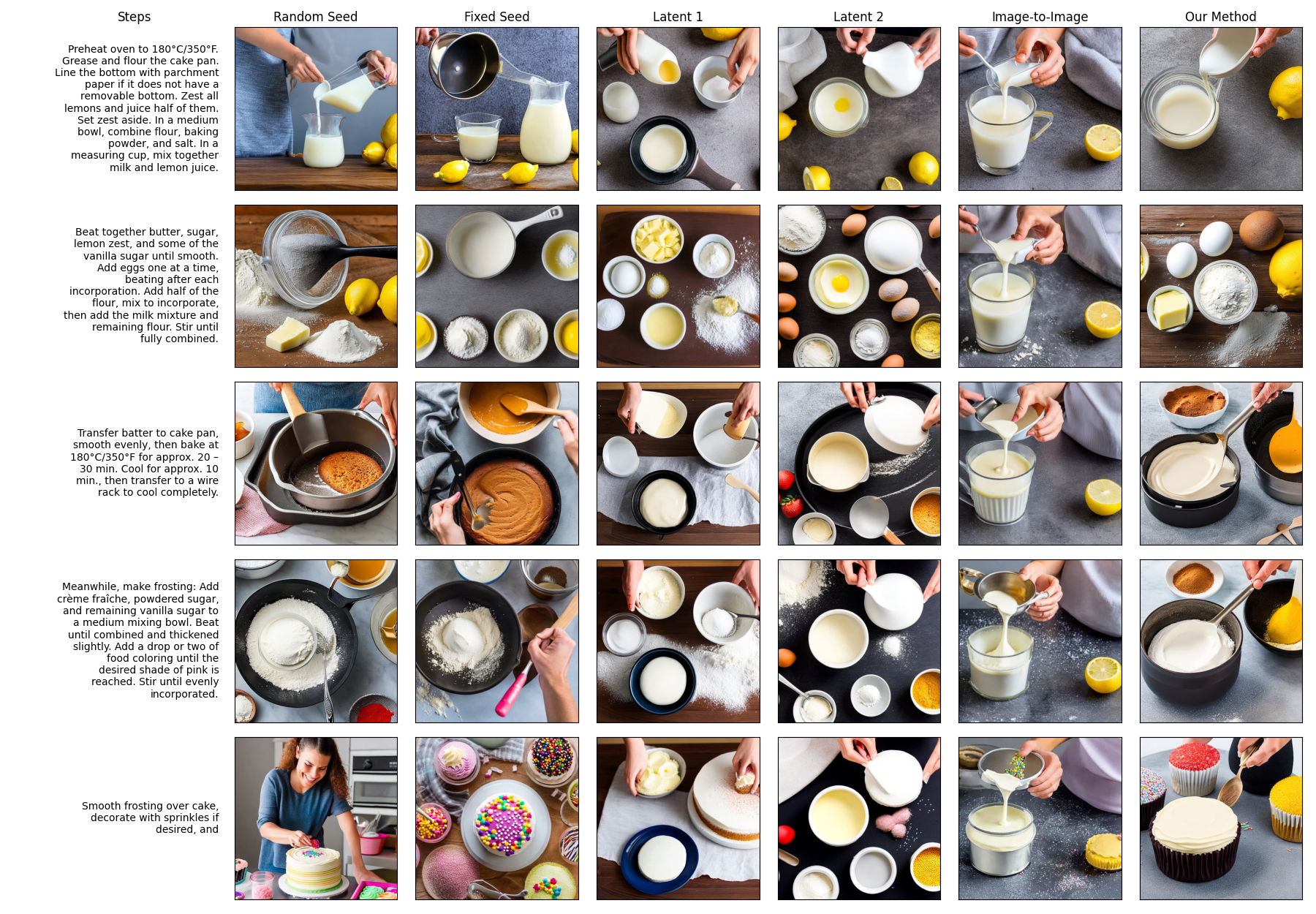}
  \caption{Examples of recipe illustrations with different methods for maintaining visual coherence.}
  \label{fig:visual-coherence-recipe-10}
\end{figure*}

\begin{figure*}[t]
  \centering
    \includegraphics[width=1.0\linewidth]{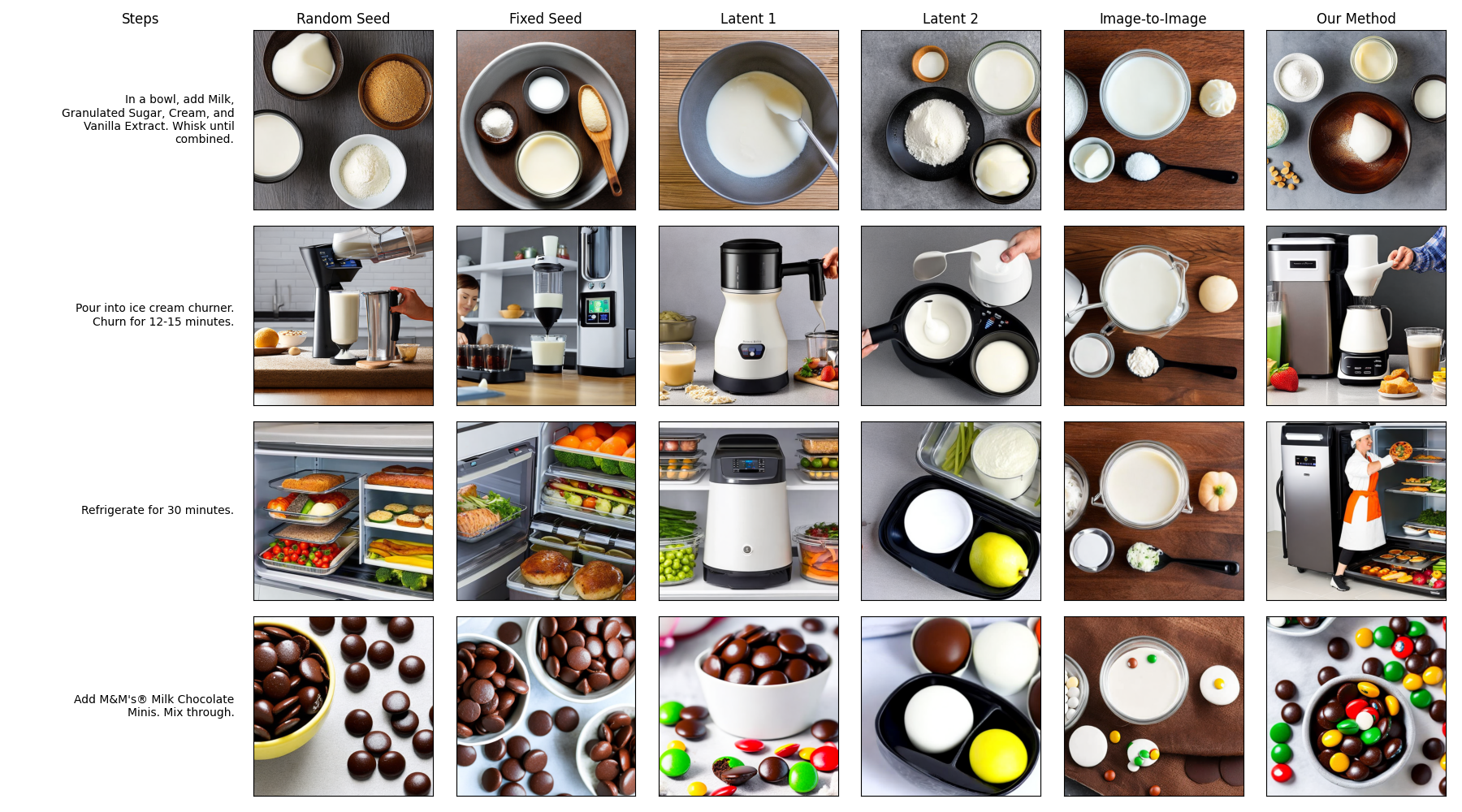}
  \caption{Examples of recipe illustrations with different methods for maintaining visual coherence.}
  \label{fig:visual-coherence-recipe-15}
\end{figure*}

\begin{figure*}[t]
  \centering
    \includegraphics[width=1.0\linewidth]{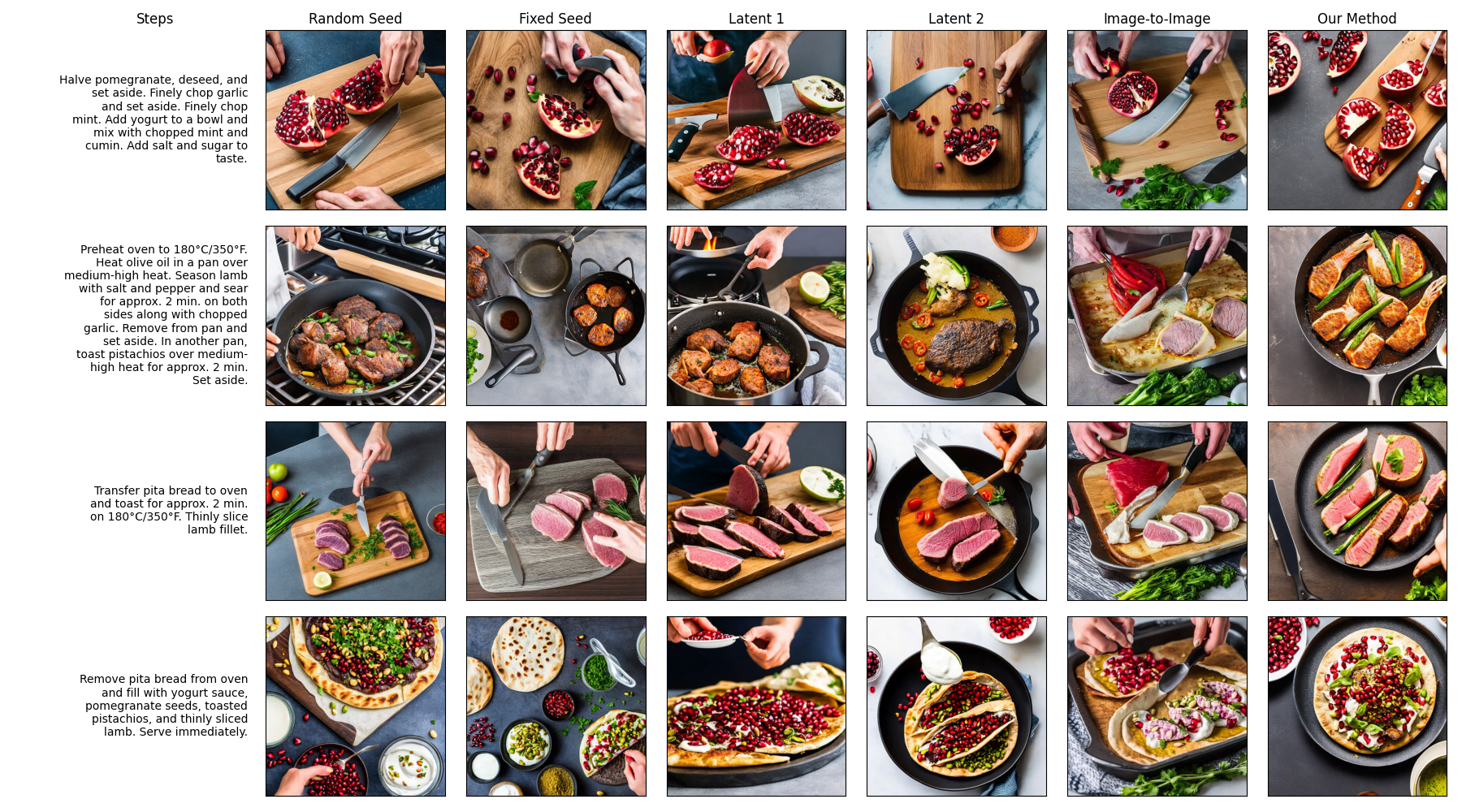}
  \caption{Examples of recipe illustrations with different methods for maintaining visual coherence.}
  \label{fig:visual-coherence-recipe-24}
\end{figure*}

\begin{figure*}[t]
  \centering
    \includegraphics[width=1.0\linewidth]{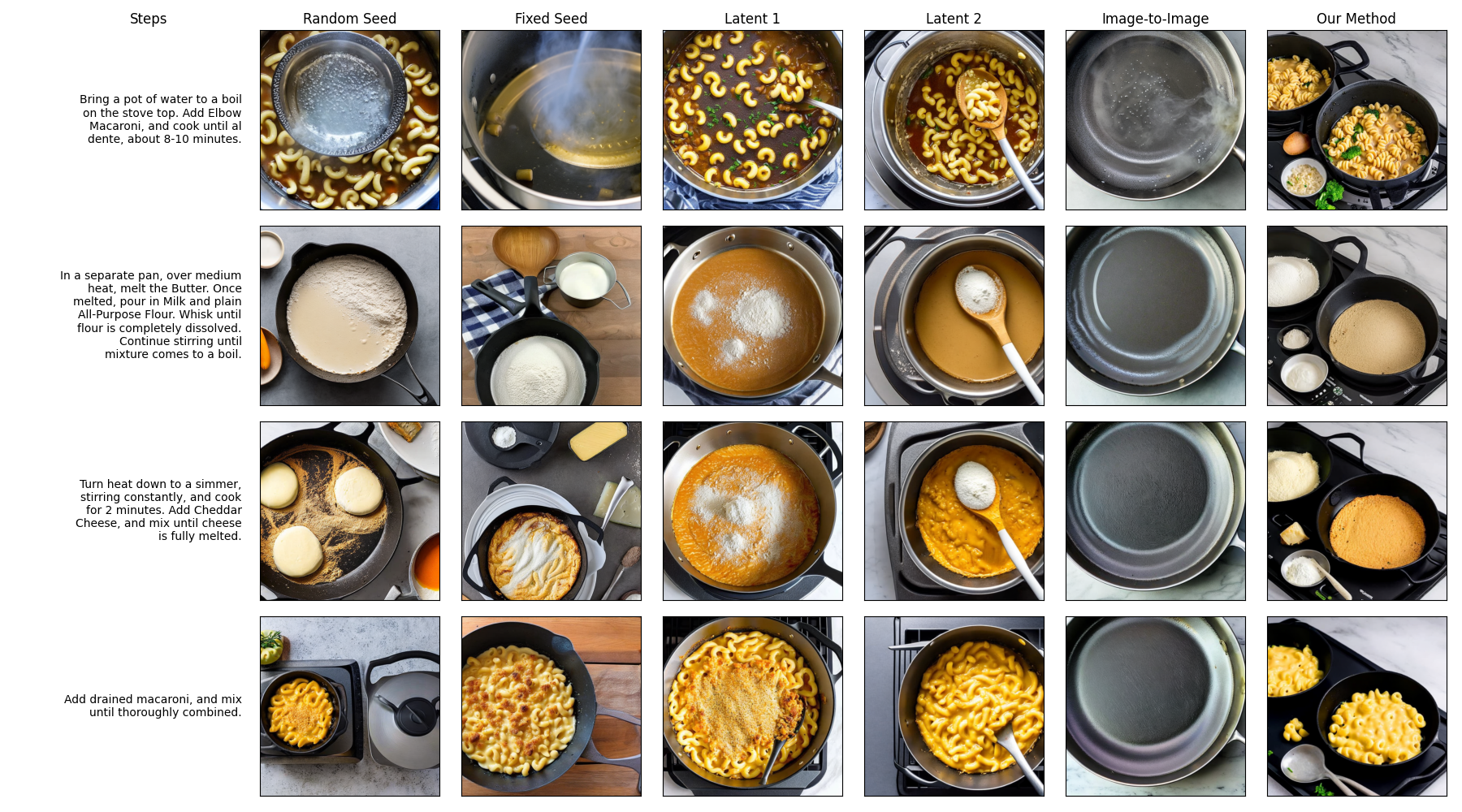}
  \caption{Examples of recipe illustrations with different methods for maintaining visual coherence.}
  \label{fig:visual-coherence-recipe-28}
\end{figure*}

\begin{table*}
    \centering
    \begin{tabular}{p{3cm}|p{12cm}}
        \toprule
        \multicolumn{2}{L{15cm}}{\textbf{Example 1: The Sequence Context Decoder was able to dereference the objects from caption$_{n-2}$ and create the correct caption for step$_n$.}}\\
        \midrule
        \textbf{Caption$_{n-2}$} & A wooden cutting board with ingredients for a salad including \textbf{avocado}, \textbf{crab meat}, and \textbf{grapes}\\
        \midrule
        \textbf{Step$_{n-1}$} & Dress all ingredients except the arugula with one fourth cup dressing, dress the arugula separately with two tablespoon dressing.\\
        \midrule
        \textbf{Step$_n$} & Plate the arugula first. Combine the rest of the ingredients, and add on top.\\
        \midrule
        \textbf{Output} & A plate of salad with \textbf{avocado}, \textbf{crab meat}, and \textbf{grapes} on top.\\
        \bottomrule
        \multicolumn{2}{l}{\textbf{}}\\
        \toprule
        \multicolumn{2}{L{15cm}}{\textbf{Example 2: The state of objects is not always preserved in the generated caption.}}\\
        \midrule
        \textbf{Caption$_{n-1}$} & A purple plastic cup is sitting on a wooden table. Inside the cup, there is a mixture of black pepper, olive oil, brown sugar, curry powder, salt, and white vinegar. There are also some sultanas floating in the mixture. The cup is covered with plastic wrap.\\
        \midrule
        \textbf{Step$_n$} & Get White Rice in a big bowl. Add in Celery and Scallion. In order to add a bit of color and some sweetness, \textbf{chop up Red Apple} and dust that up. Toss them together.\\
        \midrule
        \textbf{Output} & A bowl of white rice is being prepared with celery, scallions, and \textbf{red apples}\\
        \bottomrule
        \multicolumn{2}{l}{\textbf{}}\\
        \toprule
        \multicolumn{2}{l}{\textbf{Example 3: Some objects are correctly added to the generated caption.}}\\
        \midrule
        \textbf{Caption$_{n-1}$} & In the image, there is a pot of soup on the stove with chicken and vegetables in it. There is also a person's hand reaching into the pot to add more ingredients to the soup.\\
        \midrule
        \textbf{Step$_n$} & Serve with White Rice on the bottom and Lime Wedge on the side. Or with tortillas for dipping into the soup.\\
        \midrule
        \textbf{Output} & In the image, there is a \textbf{bowl of soup} with a tortilla in it. The soup has chicken, vegetables, and \textbf{lime juice} in it.\\
        \bottomrule
        \multicolumn{2}{l}{\textbf{}}\\
        \toprule
        \multicolumn{2}{L{15cm}}{\textbf{Example 4: Sequence of actions was not fully captured by the model.}}\\
        \midrule
        \textbf{Step$_{n-1}$} & Bring Water to boil in a medium saucepan.\\
        \midrule
        \textbf{Step$_n$} & Once water has boiled, remove saucepan from heat and add the Black Tea. Steep for 15 minutes.\\
        \midrule
        \textbf{Output} & a saucepan with black tea in it on a stove top\\
        \bottomrule
        \multicolumn{2}{l}{\textbf{}}\\
        \toprule
        \multicolumn{2}{L{15cm}}{\textbf{Example 5: When steps provide plenty of context, using a single step is sufficient.}}\\
        \midrule
        \textbf{Step $n$} & Halve fennel, remove stalk, and cut into fine strips. Finely dice onion. Peel garlic. Clean zucchini and bell pepper and cut into bite-sized pieces. Combine vegetables with rosemary, thyme, and some of the olive oil in a plastic bag. Close, shake, and marinate for approx. 20 min.\\
        \midrule
        \textbf{Output} & A woman is cutting up vegetables on a cutting board with a knife and a peeler.\\ 
        \bottomrule
    \end{tabular}
    \caption{Qualitative analysis of the Sequence Context Decoder results.}
    \label{tab:llm-generation-examples}
\end{table*}

\begin{figure*}[t]
  \centering
    \includegraphics[width=1.0\linewidth]{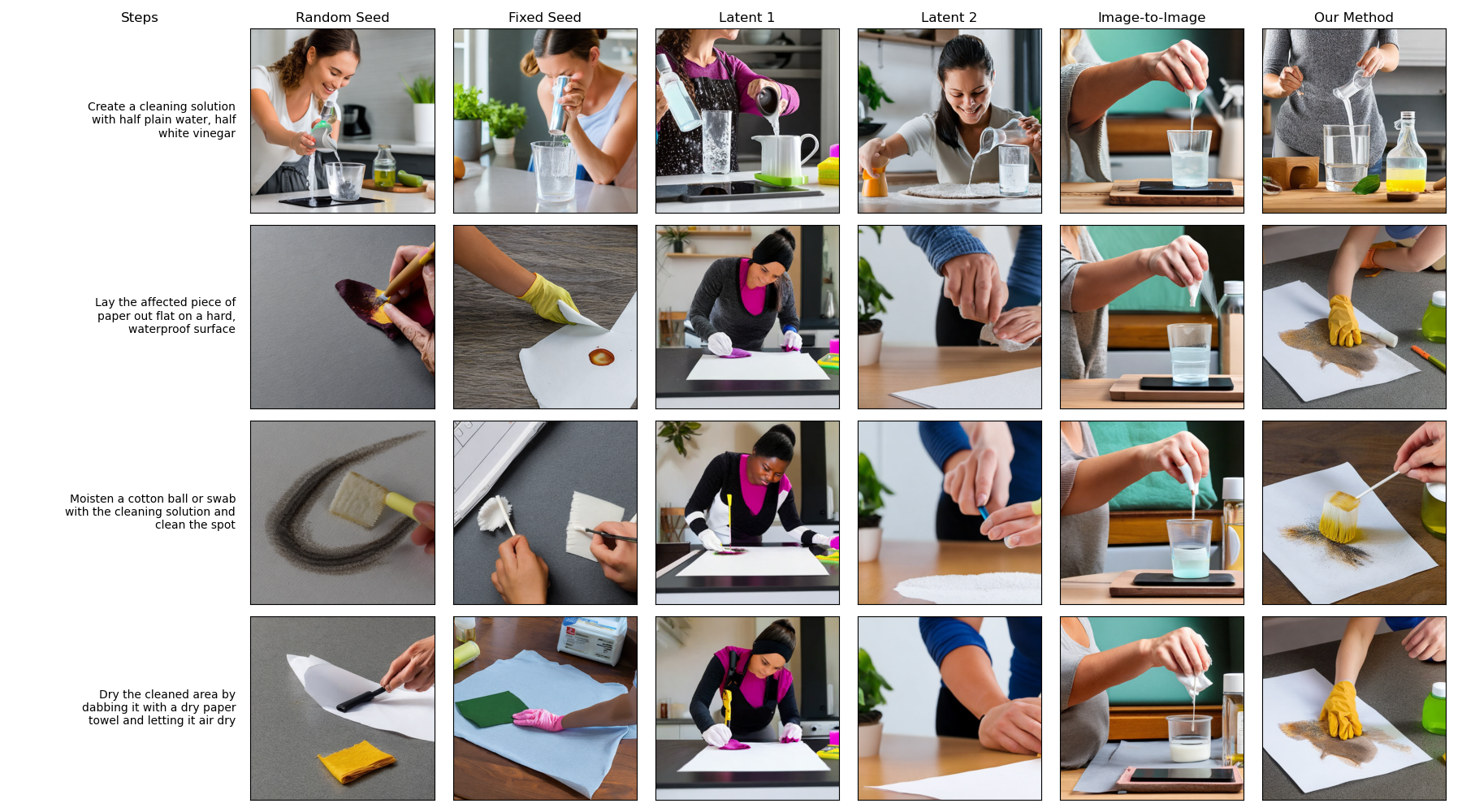}
  \caption{Examples of task illustrations with different methods for maintaining visual coherence.}
  \label{fig:wikihow-clean-surface}
\end{figure*}

\begin{figure*}[t]
  \centering
    \includegraphics[width=1.0\linewidth]{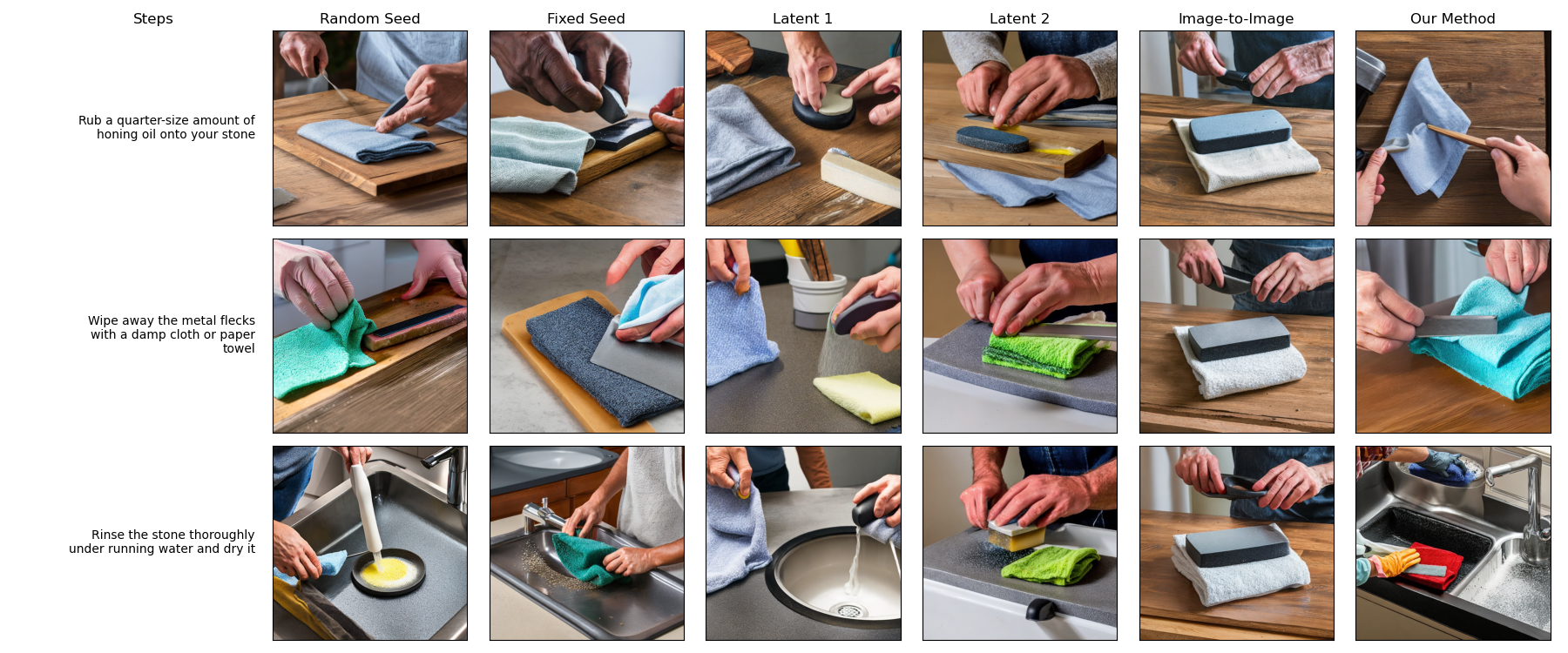}
  \caption{Examples of task illustrations with different methods for maintaining visual coherence.}
  \label{fig:wikihow-clean-stone}
\end{figure*}

\begin{figure*}[t]
  \centering
    \includegraphics[width=1.0\linewidth]{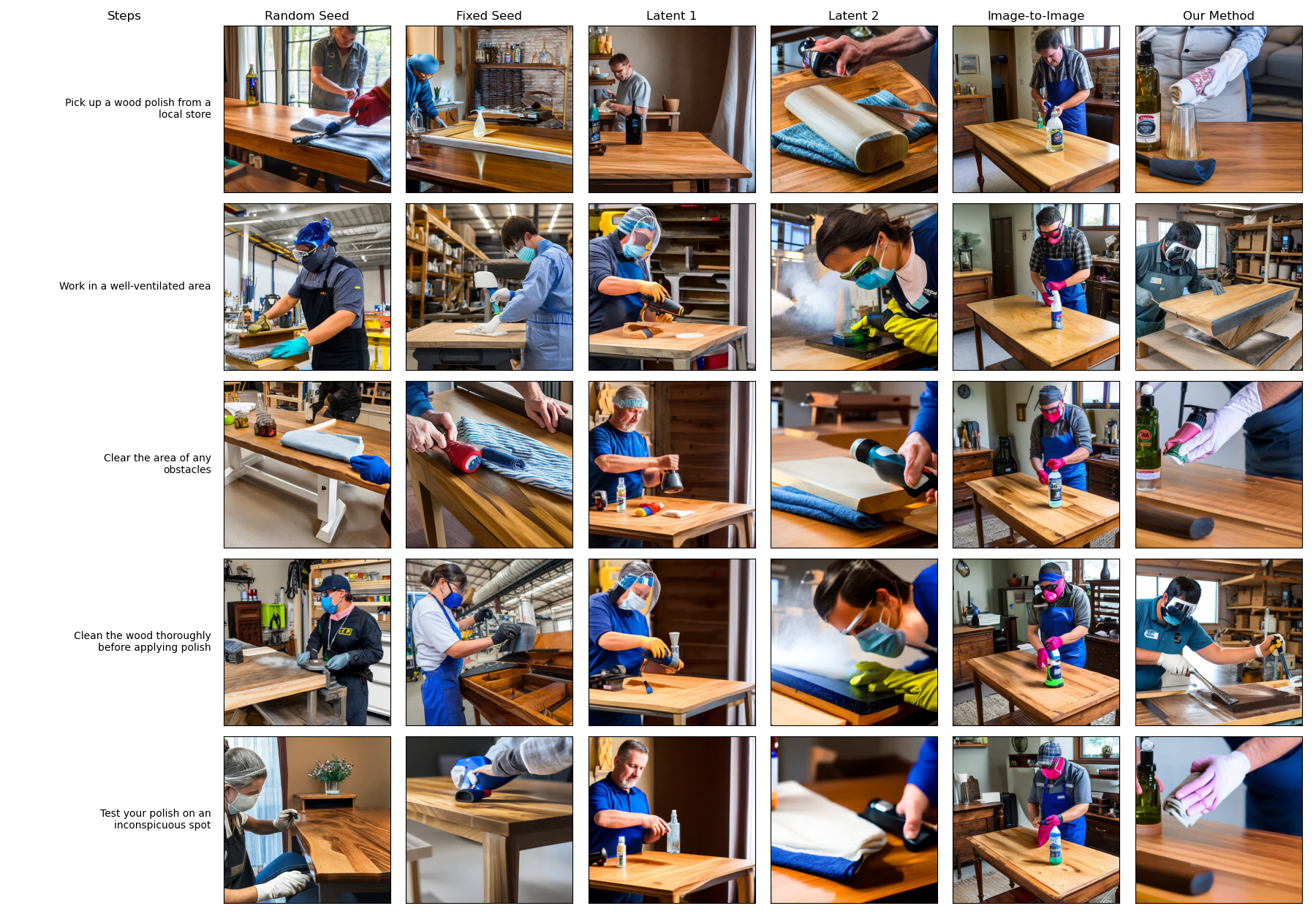}
  \caption{Examples of task illustrations with different methods for maintaining visual coherence.}
  \label{fig:wikihow-wood}
\end{figure*}

\begin{figure*}[t]
  \centering
    \includegraphics[width=1.0\linewidth]{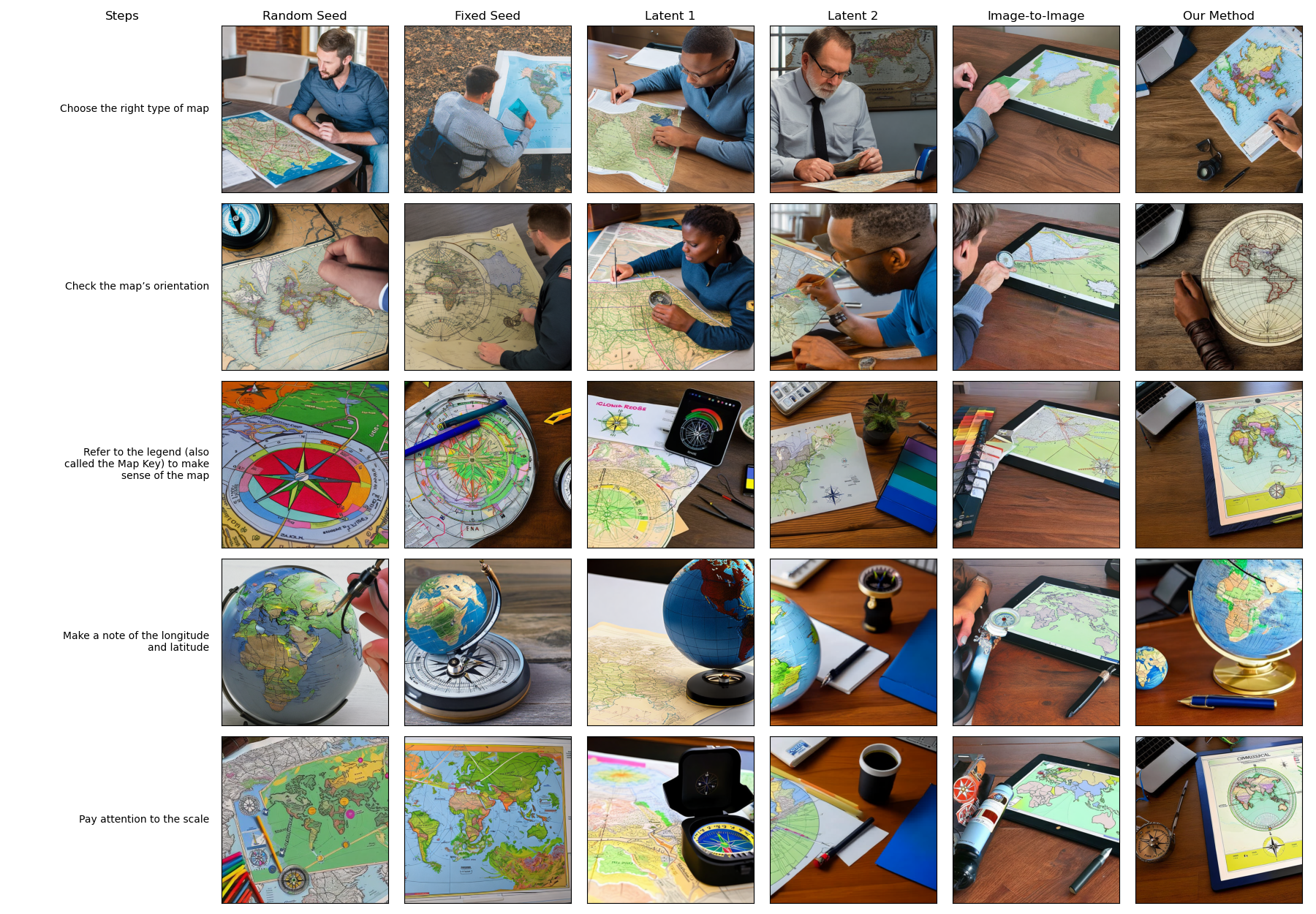}
  \caption{Example of a task that is very challenging to illustrate. We can see how the generated images still capture some of the more challenging elements of the steps, such as "Make a note of the longitude and latitude" in step 4, with the images showing a pen.}
  \label{fig:wikihow-map-task}
\end{figure*}

\end{document}